\documentclass{article}

\usepackage{PRIMEarxiv}
\usepackage[table]{xcolor}
\usepackage[utf8]{inputenc} 
\usepackage[T1]{fontenc}    
\usepackage{hyperref}       
\usepackage{url}            
\usepackage{booktabs}       
\usepackage{amsfonts}       
\usepackage{nicefrac}       
\usepackage{microtype}      
\usepackage{lipsum}
\usepackage{fancyhdr}       
\usepackage{graphicx}       
\graphicspath{{media/}}     
\usepackage{natbib}
\usepackage{amsmath}
\usepackage{subcaption}
\usepackage{wrapfig}

\newcommand{\namedpar}[1]{\vspace{0.1cm}\textbf{#1}.}

\title{Reinforcement Learning for Durable \\Algorithmic Recourse}

\author{
  Marina Ceccon \\
  University of Padova\\
  Padova \\
  \texttt{marina.ceccon@phd.unipd.it} \\
   \And
  Alessandro Fabris \\
  Univesity of Trieste \\
  Trieste \\
  \texttt{alessandro.fabris@units.it} \\
  \And
  Goran Radanović \\
  Max Planck Institute for Software Systems \\
  Saarbrücken \\
  \texttt{gradanovic@mpi-sws.org} \\
    \And
  Asia J. Biega \\
  Max Planck Institute for Security and Privacy \\
  Bochum \\
  \texttt{asia.biega@mpi-sp.org} \\
      \And
  Gian Antonio Susto \\
  University of Padova \\
  Padova \\
  \texttt{gianantonio.susto@unipd.it} \\
}

\begin{document}
\maketitle

\begin{abstract}
Algorithmic recourse seeks to provide individuals with actionable recommendations that increase their chances of receiving favorable outcomes from automated decision systems (e.g., loan approvals). While prior research has emphasized robustness to model updates, considerably less attention has been given to the \emph{temporal dynamics} of recourse---particularly in competitive, resource-constrained settings where recommendations shape future applicant pools. In this work, we present a novel time-aware framework for algorithmic recourse, explicitly modeling how candidate populations adapt in response to recommendations. Additionally, we introduce a novel reinforcement learning (RL)-based recourse algorithm that captures the evolving dynamics of the environment to generate recommendations that are both feasible and valid.
We design our recommendations to be \emph{durable}, supporting validity over a predefined time horizon $T$. This durability allows individuals to confidently reapply after taking time to implement the suggested changes. Through extensive experiments in complex simulation environments, we show that our approach substantially outperforms existing baselines, offering a superior balance between feasibility and long-term validity. 
Together, these results underscore the importance of incorporating temporal and behavioral dynamics into the design of practical recourse systems. 
\end{abstract}


\section{Introduction}
Algorithmic recourse seeks to provide individuals who have been rejected by automated decision-making systems with counterfactual explanations that clarify the reasons for their rejection~\citep{karimi2022survey, rasouli2024care, rawal2020beyond}. These explanations typically consist of alternative feature values, close to the original ones, that would have led to a favorable decision~\citep{ wachter2018counterfactualexplanationsopeningblack, barocas2020hidden}. 

Actionable recommendations based on counterfactual explanations enable individuals to improve their chances of acceptance in the future \citep{karimi2020algorithmicrecoursecounterfactualexplanations, upadhyay2025counterfactual}. However, \emph{shifts in the training data, prediction model, or applicant pool can render such recommendations invalid over time, leading to situations where individuals who follow the suggested changes---often at significant time, labor, or financial costs---still get rejected}~\citep{upadhyay2021towards, fonseca2023setting}.
\begin{figure*}[t]
    \centering
        \includegraphics[width=\linewidth]{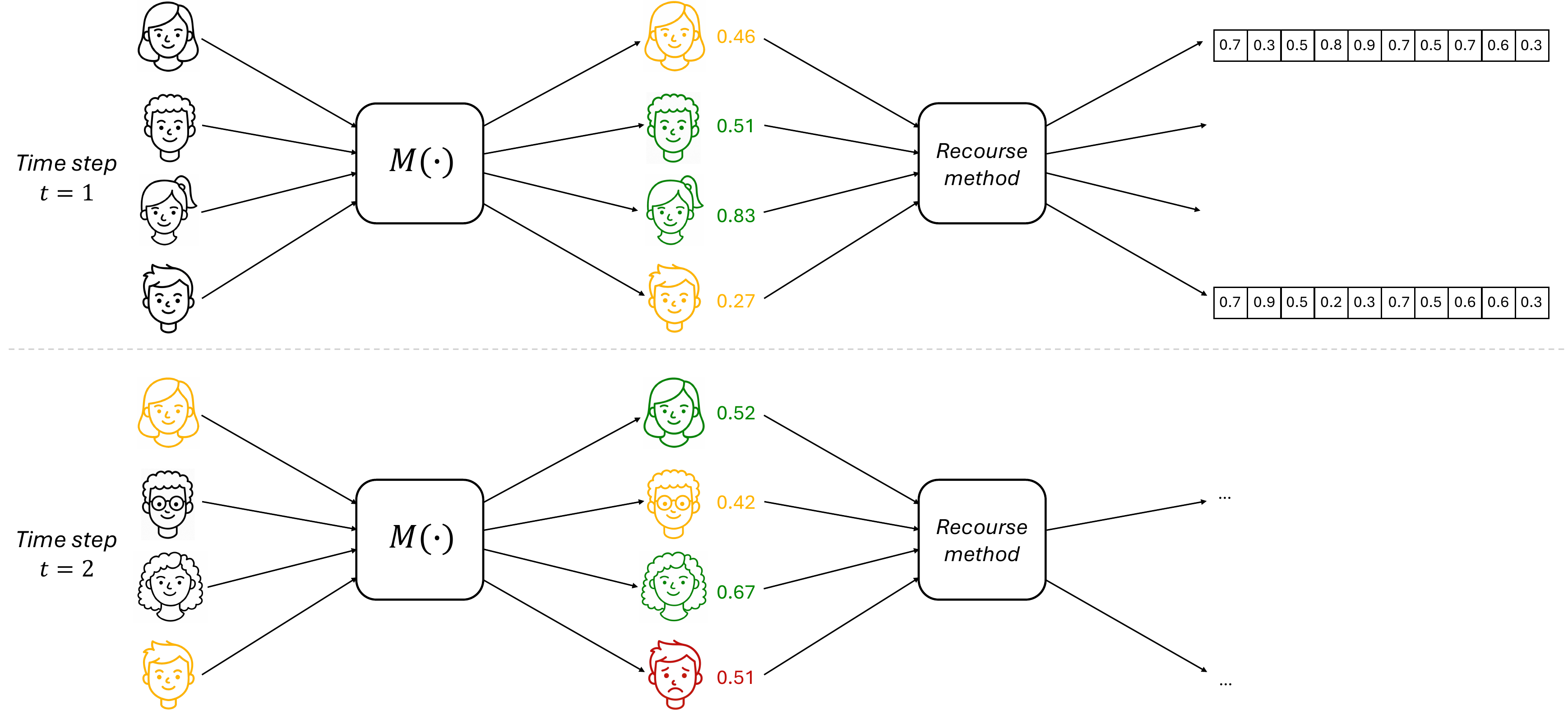}
    \caption{
    \emph{Recourse invalidity}. 
    At $t=1$, four candidates apply, and the two with the highest scores are accepted. 
    The decision threshold is $0.51$; following the state-of-the-art approach, rejected candidates (yellow) receive recommendations to reach this score. 
    At $t=2$, the rejected candidates from $t=1$ (yellow) reapply, along with two new candidates (black). 
    The yellow candidates have implemented the recommendations and raised their scores to around $0.51$. However, because of simultaneous recourse and a new candidate with a higher score, one reapplicant is still rejected.}
    \label{fig:RecourseSetting}
\end{figure*}
This issue of unreliable recourse is critical to address as it undermines trust in the system, may discourage individuals from engaging with it, and result in wasted effort~\citep{rawal2020algorithmic}. 


This concern has motivated the development of robust recourse methods that seek to remain effective in dynamic settings, contingent on the socio-technical context in which the system operates and responsive to the evolving conditions of the decision-making system and its environment~\citep{upadhyay2021towards, dominguez2022adversarial, pawelczyk2023probabilistically}. In particular, when considering \textit{limited-resource, competitive} settings, it becomes essential to account for and manage the feedback effects of recourse on the applicant pool~\citep{fonseca2023setting}. Namely, as candidates repeatedly apply after attempting to follow the recommendations, the decision threshold may shift, potentially leading to a high rate of invalidity~\citep{bell2024game}; we exemplify this challenge in Appendix~\ref{app:example} with a motivating scenario. While prior work has identified this issue and emphasized the limitations of existing recourse methods under such endogenous dynamics~\citep{fonseca2023setting, segal2024better}, \emph{no comprehensive solution has yet been proposed}. 




In this work, we address this gap by modeling the problem through the lens of reinforcement learning (RL), interpreting the recommendation process as the policy of an RL agent, thereby capturing the sequential nature of interactions between the system and the applicants. The agent is trained to provide recommendations that are feasible, robust, and valid over a predefined time horizon $T$.
Our contributions are as follows:
\begin{itemize}
    \item We introduce a comprehensive, time-aware recourse framework that models a competitive, limited-resource setting in which recommendations are issued. Our environment captures feature-modification difficulties and delays between candidate reapplications, reflecting complex human behavior and contextual constraints.
    \item We propose a novel RL-based recourse algorithm that explicitly accounts for the feedback effects of recommendations on the applicant pool. To our knowledge, this is the first solution to the challenge of providing recourse in dynamic, resource-constrained environments. Our recommendations come with guarantees of validity over a configurable time horizon $T$, allowing candidates to delay reapplication while still benefiting from the same guidance.
    \item Through extensive experiments, we demonstrate the superiority of our method over the state-of-the-art, and we analyze how intrinsic context characteristics and stakeholder objectives shape the trade-off between the feasibility and validity of recommendations.
\end{itemize}
\section{Related work}
\namedpar{Foundational Works on Algorithmic Recourse} Algorithmic recourse emerged in response to concerns about the opacity of automated decision-making, particularly in the context of the GDPR's \emph{right to an explanation}. A foundational contribution came from~\citet{wachter2018counterfactualexplanationsopeningblack}, who introduced counterfactual explanations as a way to help individuals understand and contest model outcomes; they casted recourse as an optimization problem, where the goal was identifying the smallest set of feature changes that would alter a decision. Building on this idea,~\citet{ustun2019actionable} formalized recourse in terms of practical costs, proposing integer programming methods to generate actionable changes for linear classifiers. Subsequent work generalized these approaches to incorporate richer objectives~\citep{dandl2020multi, mothilal2020explaining, cheon2024feature, rasouli2024care, rawal2020beyond}. Among these,~\citet{mothilal2020explaining} 
introduced DiCE, which generates diverse sets of feasible counterfactuals.

A parallel line of research situates recourse within the framework of structural causal models (SCMs), emphasizing that feature dependencies constrain which interventions are feasible and meaningful~\citep{karimi2020algorithmicrecoursecounterfactualexplanations, beretta2023importance}. Early work assumed access to the true underlying SCM~\citep{karimi2020algorithmicrecoursecounterfactualexplanations}, whereas more recent methods seek to approximate the causal structure in practice~\citep{karimi2020algorithmic, majumdar2024carma}.

\namedpar{Dynamic recourse (exogenous shifts)} Another strand of work examines the robustness of recourse in dynamic environments~\citep{Altmeyer_2023, yang2025robustmodelevolutionalgorithmic, kayastha2024learning, stkepka2025counterfactual, de2025time}. Existing work has largely focused on \emph{exogenous model shifts in non-competitive settings}~\citep{upadhyay2021towards, pawelczyk2023probabilistically, guyomard2023generating, nguyen2023distributionally}. \citet{upadhyay2021towards} proposed a min–max optimization framework that ensures recourse validity under worst-case perturbations to model parameters and inputs. \citet{dominguez2022adversarial} introduced adversarially robust strategies for counterfactual generation, while~\citet{pawelczyk2022trade} highlighted the trade-off between robustness and compliance with the right to be forgotten.

\namedpar{Reinforcement Learning solutions} Recent research incorporates reinforcement learning into algorithmic recourse. For instance,~\citet{de2022personalized} leverage RL to learn individual preferences and generate tailored recourse plans. Going further,~\citet{kanamori2025algorithmic} apply RL to the concept of \emph{improvement}~\citep{konig2023improvement}, ensuring recommendations not only increase the chance of acceptance but also positively affect the system where the recourse is issued.
Other work highlights the role of \emph{risk} and imperfect user execution:~\citet{wu2024safear} use RL to balance cost and risk, providing policies that allow individuals to select safer options, while~\citet{xuan2025perfect} use RL to generate robust action trajectories that account for imperfect execution. 

\namedpar{Competitive, limited-resourse setting} \citet{fonseca2023setting} and~\citet{bell2024game} explore \emph{endogenous population shifts} induced by recourse in \emph{competitive environments}. They introduce an agent-based simulation framework to analyze how applicant competition affects recourse validity. They conclude that the state-of-the-art approach of pushing rejected candidates towards the last-seen decision threshold is ineffective, as it leads to high values of invalidity. 
Although these works highlight the challenge of maintaining valid recourse under competition, they stop short of offering concrete solutions.

While prior work has significantly contributed to the field of algorithmic recourse, existing approaches primarily focus on improving individual recommendations. The literature, however, largely overlooks the endogenous feedback dynamics that arise in competitive environments with multiple candidates, where limited resources and strategic interactions continually reshape the decision boundary. This work addresses this critical gap, proposing a novel reinforcement learning method to generate feasible recourse recommendations that remain valid over a finite time horizon.


\section{Competitive Recourse Setting}
In this section, we introduce the setting of the problem under study. We first describe the simulation environment in which candidates compete for a limited resource and modify their features based on recourse recommendations. We then formalize this environment as a reinforcement learning problem, where the objective is to identify an optimal policy for generating recommendations.

\subsection{Simulation Environment}
\label{sec:simulation}

We build our time-aware recourse framework on prior work modeling recourse under limited resources and repeated applications~\citep{fonseca2023setting,bell2024game}, while introducing additional mechanisms to more thoroughly capture the dynamics of competitive recourse systems.

The simulation begins with an initial population $\mathcal{I}_0$ of $N_0$ candidates. More generally, we denote the population at time $t$ by $\mathcal{I}_t$, with size $N_t$. Each candidate $j$ is characterized by a feature vector $X_0^{\mathrm{F}}[j] \in [0,1]^z$, where $X_t^{\mathrm{F}} \in [0,1]^{N_t \times z}$ denotes the matrix of \emph{factual} features for the candidate pool at time step $t$, and $z$ is the total number of features, that take values in $[0,1]$. Following prior work on competitive recourse~\citep{fonseca2023setting,bell2024game}, we assume that features are independently sampled from their respective marginal distributions, without causal dependencies among them. The details of the synthetic feature generation process are provided in Appendix~\ref{app:environment}.


At each time step $t = 0,1,2,\dots$, the population evolves as $m$ new candidates enter, $k$ candidates are accepted, and a variable number of candidates leave. A previously trained prediction model $M : [0,1]^z \to [0,1]$ assigns a qualification score to each feature vector. At each step, a threshold $th_t$ is chosen so that exactly $k$ candidates in $\mathcal{I}_t$ are accepted. For a candidate $j \in \{1,\dots,N_t\}$ with features $X_t^{\mathrm{F}}[j]$, the acceptance indicator is defined as
$
h_k\big(M(X_t^{\mathrm{F}}[j]), th_t\big) \in \{0,1\},
$
where $1$ denotes acceptance and $0$ rejection.

Each rejected candidate $j$ is offered recourse in the form of a \emph{counterfactual} feature vector $X^{\mathrm{CF}}[j]$, designed to ensure acceptance within $T$ time steps. Candidates decide whether to attempt the modification or exit the environment. This decision is governed by a \emph{dropout probability}, which increases with both the number of failed attempts and the magnitude of required changes, modeling candidate discouragement~\citep{grbic2012factors}. 

For candidates who remain, each modification on each feature $i$ is implemented successfully with a \emph{probability of success} that depends on the change magnitude, a feature-specific difficulty parameter $d_i \in [0,1]$~\citep{lievens2005retest}, and a global difficulty parameter $\beta$. In addition, each candidate has a \emph{reapplication probability}, which increases with (i) \emph{self-confidence}, measured by the extent to which recommended changes were applied, and (ii) \emph{urgency}, determined by the time since the last application~\citep{grbic2012factors}. Candidates may delay reapplication for up to $T$ steps, consistent with the guaranteed validity of the recommendation.
These extensions improve upon prior simulations~\citep{fonseca2023setting, bell2024game}, which assumed (i) zero dropout probability, meaning that candidates only left once accepted, (ii) uniform modification difficulty across features, and (iii) immediate reapplication without guarantees on recommendation duration. A detailed specification of the mechanisms here presented and the improvements with respect to previous work is provided in Appendix~\ref{app:environment}. 
\subsection{Reinforcement learning setting}
\label{sec:simulation2}
We model the environment where the reinforcement learning agent is trained as a \textit{Partially Observable Markov Decision Process (POMDP)}, capturing the sequential nature of algorithmic recourse under feedback loops, and extending the simulation framework introduced earlier.


Partial observability arises due to delays in candidate reapplications and exits. Individuals modify their features in response to prior recommendations, but these changes remain hidden until they reapply, if they do at all. Some may permanently exit the system due to discouragement, introducing further uncertainty into the environment.

Formally, the environment is a POMDP specified by the tuple $(\mathcal{S}, \mathcal{A}, \mathcal{P}, \Omega, O, R, \gamma)$, where $\mathcal{S}$ is the latent state space, $\mathcal{A}$ the set of actions, and $\mathcal{P}(s'|s, a)$ the transition function that defines the probability of moving from state $s$ to $s'$ after taking action $a$. The agent receives partial observations from an observation space $\Omega$, governed by the observation function $O(o|s', a)$, which defines the likelihood of observing $o$ upon reaching $s'$ via action $a$. The reward function $R(s, a)$ assigns a scalar signal to each state-action pair, and $\gamma \in [0,1]$ is a discount factor balancing immediate and future rewards. We now describe the main components of the POMDP, starting from the latent state.

\namedpar{State $s_t$} The state $s_t$ captures the complete configuration of the environment at time $t$. It includes all candidates currently in the system, represented by their feature matrix $X_{c,t}$ and identifiers $\mathcal{I}_{c,t}$, as well as all candidates applying at this step—including new entrants—represented by $X_t^{\mathrm{F}}$ and $\mathcal{I}_t^{\mathrm{F}}$. Scores and binary outcomes for all candidates are obtained via the decision model $M(\cdot)$ and the acceptance indicator $h(\cdot)$. The state space $\mathcal{S}$ is continuous---since candidate features and scores are continuous---and its dimension varies with the number of candidates present and those reapplying.

\namedpar{Action $a_t$} The agent's action at time $t$ is defined as: $a_t = X_{t}^{\text{CF}}$, where $X_{t}^{\text{CF}}$ is a matrix of counterfactual feature vectors, each corresponding to a rejected candidate. These vectors represent the configurations that, if adopted, would lead to acceptance, within a time window of $T$ steps. The action space $\mathcal{A}$ is continuous and of variable dimension.

\namedpar{Transition Function $\mathcal{P}(s_{t+1}|s_t, a_t)$} The environment evolves according to a stochastic transition function $\mathcal{P}$, mapping the current state $s_t$ and agent action $a_t$ to a distribution over successor states $s_{t+1}$. 
Transitions proceed in three phases. First, candidates with positive outcomes permanently exit the environment and are removed from $X_{c,t+1}$. Second, rejected candidates respond to their counterfactual recourse recommendations: some exit due to discouragement, while others remain and modify their features toward the suggested counterfactuals, updating $X_{c,t+1}$. Finally, a new application round occurs, comprising both new entrants and reapplying candidates previously rejected, forming the new feature matrix $X_{t+1}^{\mathrm{F}}$.



\namedpar{Observation $o_t$ and Observation Function $O(o_t \mid s_t, a_{t-1})$} 
The agent has partial observability of the environment, and the observation function specifies how this partial view is derived from the true state $s_t$ and the action $a_{t-1}$. Formally, the observation includes only the elements of $s_t$ corresponding to the current applicants: 
$
o_t = (X_{t}^{\text{F}}, \mathcal{I}_{t}^{\text{F}}).
$
The observation space $\Omega$ is continuous and has variable dimension depending on the number of applicants at time $t$.

\namedpar{Reward Function $R(s_t, a_t)$} The reward function integrates multiple objectives to ensure \textit{equity}, \textit{validity}, and \textit{feasibility} of the agent's recommendations.
To promote \textit{equity}, we minimize disparities in the scores that rejected candidates would obtain if they implemented the recommended actions. Formally, we define the set of rejected candidates at time $t$ as $\mathcal{I}_{t}^{\text{rej}}$. For each candidate $j \in \mathcal{I}_{t}^{\text{rej}}$, the \emph{goal score} is $g_{t}[j] = M(X_{t}^{\text{CF}}[j])$,
 i.e., the score the candidate would achieve if they perfectly implemented the recommendation. We argue that these scores should be similar across rejected candidates, in order to prevent unequal treatments. 
 To guarantee this, we minimize the \textit{Gini index}:
\begin{equation}
\label{eq:gini}
    \text{Gini}_t = \frac{ \sum_{i,j \in \mathcal{I}_{t}^{\text{rej}}} \left| g_{t}[j] - g_{t}[i] \right| }{2 n_{r,t} \sum_{i \in \mathcal{I}_{t}^{\text{rej}}} g_{t}[i]},
\end{equation} 
where $n_{r,t} = |\mathcal{I}_{t}^{\text{rej}}|$. Lower $\text{Gini}_t$ indicates greater equity.

To ensure \textit{validity}, we adopt the \textit{Recourse Reliability} ($\text{RR}_t$), first introduced by~\citet{fonseca2023setting}, which measures the portion of candidates that successfully implemented a recommendation and were accepted at each time step:
\begin{equation}
\label{eq:reliability}
   \text{RR}_{t}^T = \frac{|\mathcal{I}_{t}^{\text{succ}} \cap \mathcal{I}_{t}^{\text{acc}}|}{|\mathcal{I}_{t}^{\text{succ}}|}.
\end{equation}
where $\mathcal{I}_{t}^{\text{succ}}$ indicates the candidates that successfully implemented a recommendation and reapplied at time step $t$, and $\mathcal{I}_{t}^{\text{acc}}$ indicates the candidates accepted at step $t$. In the original formulation, $\mathcal{I}_t^{\text{succ}}$ included only candidates reapplying from the previous step. We extend this to candidates whose last application was within the past $T$ steps, so $\text{RR}_t^T$ measures reliability over a $T$-step horizon.


To prevent trivial solutions that maximize $\text{RR}_t^T$ by suggesting extremely difficult modifications, we introduce the \emph{Recourse Feasibility} ($\text{RF}_{t}^T$), which quantifies the fraction of candidates who received recommendations within the past $T$ steps and reapplied with a successful implementation at time $t$:
\begin{equation}
\label{eq:feasibility}
    \text{RF}_{t}^T = \frac{|\mathcal{I}_{t}^{\text{succ}}|}{|\mathcal{I}^{\text{rej}}_{t-T:t}|},
\end{equation}
where $\mathcal{I}^{\text{rej}}_{t-T:t}$ 
 is the set of candidates who last applied unsuccessfully in the window $[t-T, t-1]$, and thus could have reapplied at $t$, with a perfectly implemented recommendation. In this way, the metric penalizes failed implementations, delays, and discouragement-related exits.

\namedpar{Policy $\pi(a_t|s_t)$} The agent learns a policy $\pi(a_t|s_t)$ that defines a distribution over recommendations $a_t$, conditioned on the current environment state $s_t$. Learning this policy is challenging due to the high-dimensional, variable-sized state and action spaces. 
In the next section, we introduce a training framework that mitigates the computational burden associated with these large and dynamic spaces.

\section{Reinforcement learning Solution}
\label{sec:solution}

In the previous section, we interpreted the simulation environment through the lens of reinforcement learning, framing the recourse task as learning a policy that yields valid, durable, and feasible recommendations. In this section, we describe the strategy used to search for the optimal policy within this environment.

Directly learning the counterfactual matrix $X_t^{\text{CF}}$ is computationally expensive due to its high dimensionality and variable size. To address this, we adopt a hierarchical approach that separates counterfactual generation from goal selection, explicitly modeling their inter-dependency.
\namedpar{Counterfactual generation} We first learn a stochastic function
    $$
        \phi: (x_t^{\text{F}}, g) \mapsto \text{Dist}(x_t^{\text{CF}}),
    $$
    that defines a probability distribution over counterfactual feature vectors $x_t^{\text{CF}}$, conditioned on a candidate’s features $x_t^{\text{F}}$ and a target score $g$. Samples from this distribution are required to satisfy $M(x_t^{\text{CF}}) \approx g$ while minimizing a cost function that measures the discrepancy between $x_t^{\text{F}}$ and $x_t^{\text{CF}}$. In other words, $\phi$ specifies how to probabilistically modify features to achieve a desired score.

\namedpar{Goal selection policy} Given the pre-trained $\phi$, we learn a stochastic policy
    $$
        \mu: (X_t^{\text{F}}, \mathcal{I}_t^{\text{F}}) \mapsto \text{Dist}(g_t),
    $$
    that defines a probability distribution over target scores $g_t$. During training, the pre-trained $\phi$ translates sampled goal scores into actionable recommendations for each rejected candidate:
    $$
        g_t \sim \mu(X_t^{\text{F}}, \mathcal{I}_t^{\text{F}}), \quad
        X_t^{\text{CF}}[j] \sim \phi\big(X_t^{\text{F}}[j], g_t\big), \quad 
        \forall j \in \mathcal{I}_t^{\text{rej}}.
    $$
    The environment evolves according to these recommendations, making the training of $\mu$ dependent on $\phi$.
    
In this hierarchical setup, $\mu$ decides \emph{what score to aim for}, while $\phi$ determines \emph{how to modify the features} to reach that score; pre-training $\phi$ reduces the computational complexity and stabilizes the training of $\mu$. While $\mu$ is primarily responsible for the trade-off between \emph{Recourse Reliability} (Equation~\ref{eq:reliability}) and \emph{Recourse Feasibility} (Equation~\ref{eq:feasibility}) in the reward function, the Pareto efficiency of this trade-off largely depends on $\phi$, as the feasibility of a recommendation critically depends on the trajectory taken to reach the target score.
Additionally, $\phi$ indirectly optimizes the Gini index (Equation~\ref{eq:gini}), by providing recommendations that approximately lead to the same score for all candidates.

This two-step architecture mirrors state-of-the-art recourse methods, which typically fix the goal score $g_t$ at the last-seen decision threshold and optimize only $\phi$. Our approach instead learns $g_t$ adaptively, based on the behavior of the candidates. We further design $\phi$ to improve the balance between reliability and feasibility, targeting higher values of both metrics. Concretely, $\phi$ is implemented as an RL policy, the \emph{recourse recommender policy}, pre-trained with respect to the target-score policy, the \emph{predictor policy}. We next describe the training procedure for both agents.

\subsection{Training of the Recourse Recommender Policy}


%
The recourse recommender policy $\phi$ is trained in a simplified environment derived from the setting introduced earlier. The state at time $t$ is defined as $s_t = (x_t^{\text{F}}, g)$, where $x_t^{\text{F}}$ is the feature vector of a single candidate and $g$ a target score. The action is the counterfactual feature vector $a_t = x_t^{\text{CF}}$. Although the deployment setting of the policy remains unchanged, with respect to the setting presented in Section \ref{sec:simulation2}, the POMDP used for training is modified to pursue a different objective, which is achieving a predefined target score, and to focus exclusively on a single candidate, which is sufficient for the intended task. This reduction significantly lowers the computational burden.

Training proceeds over multiple episodes. 
At the start of an episode, a goal score $g$ is sampled such that $M(x_0^{\text{F}}) < g$. At each step, the agent proposes a recommendation $x_t^{\text{CF}}$, which the candidate attempts to implement. Each recommendation has validity $T=1$, meaning candidates reapply at every step. The episode terminates when $M(x_t^{\text{F}}) \ge g$ or a maximum number of steps is reached.

Recommendations are evaluated on two criteria: (i) \emph{accuracy:} how closely $M(x_t^{\text{CF}})$ approaches $g$, and (ii) \emph{cost:} the effort required to modify $x_t^{\text{F}}$ into $x_t^{\text{CF}}$. The accuracy objective ensures that the recommender can generate paths toward arbitrary targets, and that---once paired with the predictor---it leads to counterfactual scores that are consistent with goal scores, thereby improving the Gini-based reward (Equation~\ref{eq:gini}). 
Formally, accuracy is measured as:
\begin{equation}
\label{eq:error}
    e_t = |M(x_{t}^{\text{CF}}) - g|.
\end{equation}
The cost objective encourages minimal-effort modifications. We define an estimated cost function that penalizes large changes and prioritizes easier-to-modify features, based on estimated difficulties:
\begin{equation}
\label{eq:cost_simplified}
    \hat{c}_t = \sum_{i=1}^z |x_{t}^{\text{CF},(i)} - x_{t}^{\text{F},(i)}| \cdot \hat{d}_i,
\end{equation}
where $z$ is the number of features, and $\hat{d}_i$ is the agent’s estimate of the difficulty of modifying feature $i$. Difficulty estimates are learned adaptively; the full procedure is detailed in Appendix~\ref{app:solution}.

For optimization, we employ the Soft Actor-Critic (SAC) algorithm~\citep{haarnoja2018softactorcriticoffpolicymaximum}, a model-free, off-policy method well-suited for continuous action spaces. In our work, $\phi$ is trained \emph{online}, interacting directly with the environment; the same procedure can also be executed \emph{offline} if a sufficiently rich dataset, related to another recourse system employed in this setting and containing information on how candidates respond to recommendations, is available.


\subsection{Training of the Predictor Policy}

The predictor policy, denoted by $\mu$, is trained on the POMDP introduced in the previous section. Within the hierarchical framework, the action space is reduced to $a_t = g_t$, i.e., the selection of a target score. During training, the recourse recommender policy $\phi$ is treated as a fixed component of the environment:
it provides the counterfactual updates required to construct $X_t^{\text{CF}}$, based on $g_t$, while $\mu$ focuses solely on learning how to select appropriate goals. The reward function used to train $\mu$ excludes the Gini term, as it is entirely handled by the recourse recommender policy. 

Because the environment is only partially observable and the reward is non-Markovian, we augment both the state and observation spaces with explicit historical information. At each time step $t$, the agent receives a window of data covering all candidates who applied and were rejected during $[t-T, t-1]$. For each such candidate, the following metadata are provided: (i) their feature vector at the time of their last application, (ii) their unique identifier, (iii) the time step of their most recent application, (iv) the total number of applications they have submitted, (v) the most recent recourse recommendation received. 
By explicitly including these variables in the agent’s observation, rather than requiring it to infer or internally store past events, we ensure that the environment is fully Markovian with respect to the predictor’s decision process. This design choice facilitates stable learning in the presence of delayed effects.

Training is conducted over fixed-length episodes. At the beginning of each episode, a new population of initial applicants is generated. The predictor $\mu$ is then optimized using SAC~\citep{haarnoja2018softactorcriticoffpolicymaximum}, chosen for its sample efficiency and ability to handle continuous action spaces.

\section{Experiments}

\subsection{Setup} In this section, we present the performance evaluation of our method relative to established baselines from the literature. Additionally, we analyze how environmental constraints and design choices influence achievable performance. Our approach is compared against three widely used baselines for non-causal recourse~\citep{ustun2019actionable, wachter2018counterfactualexplanationsopeningblack, mothilal2020explaining} (hereafter called Ustun, Wachter, and DiCE). Some of these methods have also been adopted as baselines in recent studies on recourse under competition~\citep{fonseca2023setting, bell2024game}; in our framework, they serve as alternatives to the recommender $\phi$.  


Each strategy is combined with: (i) a \emph{trivial predictor}, which applies the classifier’s most recent decision threshold (reflecting standard practice in dynamic recourse), and (ii) our proposed predictor, parameterized by policy $\mu$. This yields two categories of methods: (i) \emph{baselines}, pairing each recourse strategy with the trivial predictor, and (ii) \emph{hybrids}, pairing the strategies with our predictor.

We further compare our approach with the method of~\citet{dominguez2022adversarial}, referred to as Adversarially Robust Recourse (ARR). ARR accounts for adversarial perturbations to an individual’s features; under a linear model $M(\cdot)$ with independent features, the ARR objective reduces to adjusting the target score by a margin $\varepsilon$, and then deriving counterfactuals to reach this adjusted target using the same procedure as~\citet{ustun2019actionable}.

\begin{figure*}[tb]
    \centering
    \begin{subfigure}[b]{0.45\textwidth}
        \includegraphics[width=\textwidth]{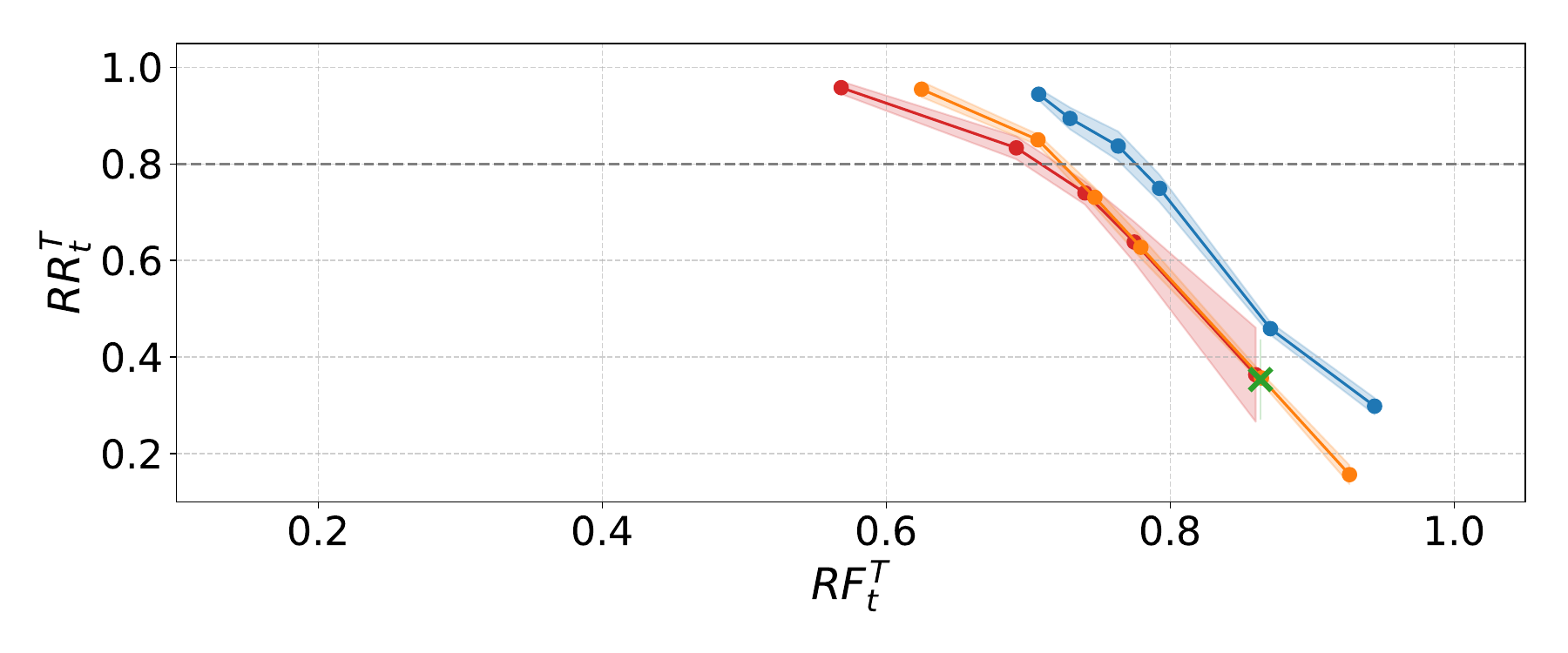}
        \caption{$T=1$, $\beta=0.05$}
        \label{fig:paretoT1EasySetting}
    \end{subfigure}
    \hfill
    \begin{subfigure}[b]{0.45\textwidth}
        \includegraphics[width=\textwidth]{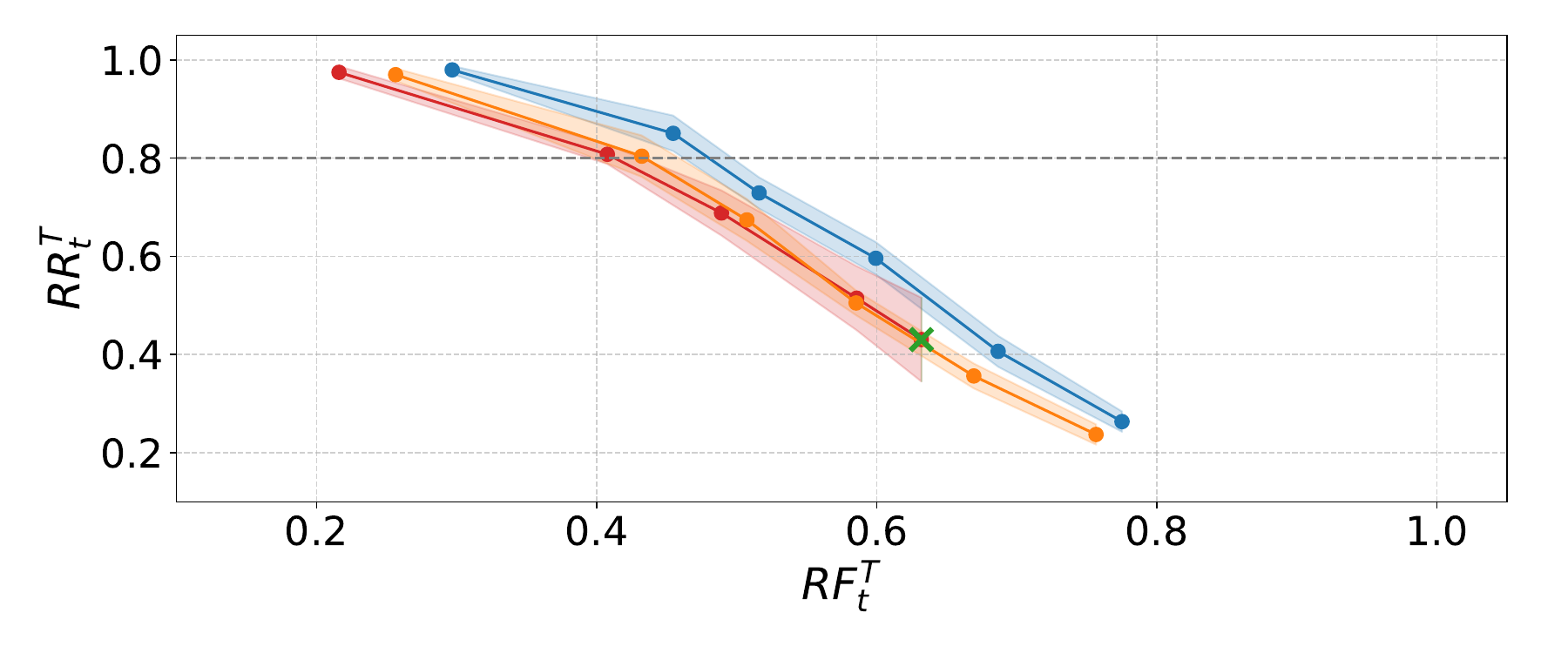}
        \caption{$T=1$, $\beta=0.01$}
        \label{fig:paretoT1}
    \end{subfigure}
    \hfill
    \begin{subfigure}[b]{0.45\textwidth}
        \includegraphics[width=\textwidth]{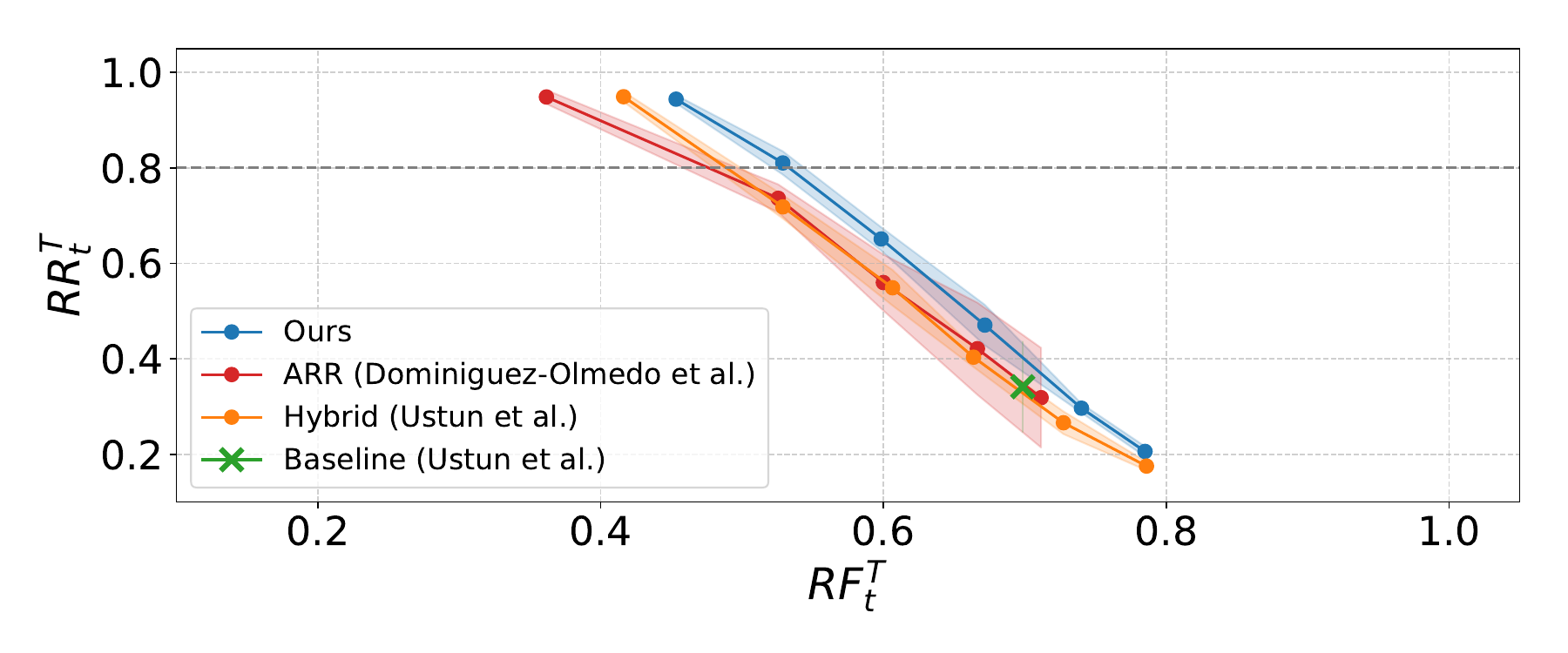}
        \caption{$T=5$, $\beta=0.05$}
        \label{fig:paretoT5EasySetting}
    \end{subfigure}
    \hfill
    \begin{subfigure}[b]{0.45\textwidth}
        \includegraphics[width=\textwidth]{paretoT=5EasySetting_Robust.pdf}
        \caption{$T=5$, $\beta=0.01$}
        \label{fig:paretoT5}
    \end{subfigure}
    
\caption{ Comparison of Pareto fronts of our method (blue), the hybrid method based on Ustun's approach (orange), the baseline using Ustun's approach (green), and the ARR method (red), across four settings with $T \in \{1,5\}$ and $\beta \in \{0.05,0.01\}$. Pareto fronts plot Recourse Reliability $\text{RR}_t^T$ and Recourse Feasibility $\text{RF}_t^T$, each averaged over ten evaluation episodes. The gray line at $RR_t=0.8$ denotes the \emph{high reliability threshold}, distinguishing configurations that achieve desirable recourse reliability.}

\label{fig:paretoGrid}
\end{figure*}

We evaluate all methods under four experimental conditions, varying the recourse horizon ($T \in \{1, 5\}$) and the setting difficulty ($\beta \in \{0.05, 0.01\}$). The reward coefficients $(\alpha, \tau)$, and the robustness coefficient $\varepsilon$, which govern the trade-off between Recourse Reliability and Recourse Feasibility, are chosen to produce Pareto frontiers spanning recourse reliability values $\text{RR}_t^T$ approximately in $(0.20, 0.95)$. 

Details on the training and evaluation procedures are provided in Appendix~\ref{app:experimental_setup}. 
For plot readability, we depict only results for Ustun and ARR in this section; analogous results---including comparisons with Wachter and DiCE, as well as an analysis of the Gini Index for all methods---are deferred to Appendix~\ref{app:results}. 
Finally, Appendix~\ref{app:german} presents results from a simulation that assesses the applicability of our method in a real-world setting, in which candidates’ initial features are drawn from the German Credit dataset.\footnote{\url{archive.ics.uci.edu/dataset/144/statlog+german+credit+data}}

\subsection{Results}

Figure~\ref{fig:paretoGrid} shows Pareto plots for Recourse Feasibility ($\text{RF}_t^T$) and Recourse Reliability ($\text{RR}_t^T$) across all four experimental settings, averaged over ten evaluation episodes. Regarding our method and the hybrid approach, each point on a Pareto front corresponds to a different trained predictor $\mu$ with varying values of the parameters $\alpha$ and $\tau$. In the case of ARR, each point corresponds to a different value of $\varepsilon$. The horizontal gray line at $RR_t = 0.8$ indicates the \emph{high reliability threshold}, highlighting simulations that achieve a desirable level of recourse reliability.

\begin{wrapfigure}{r}{0.5\linewidth}
    \centering

    \includegraphics[width=\linewidth]{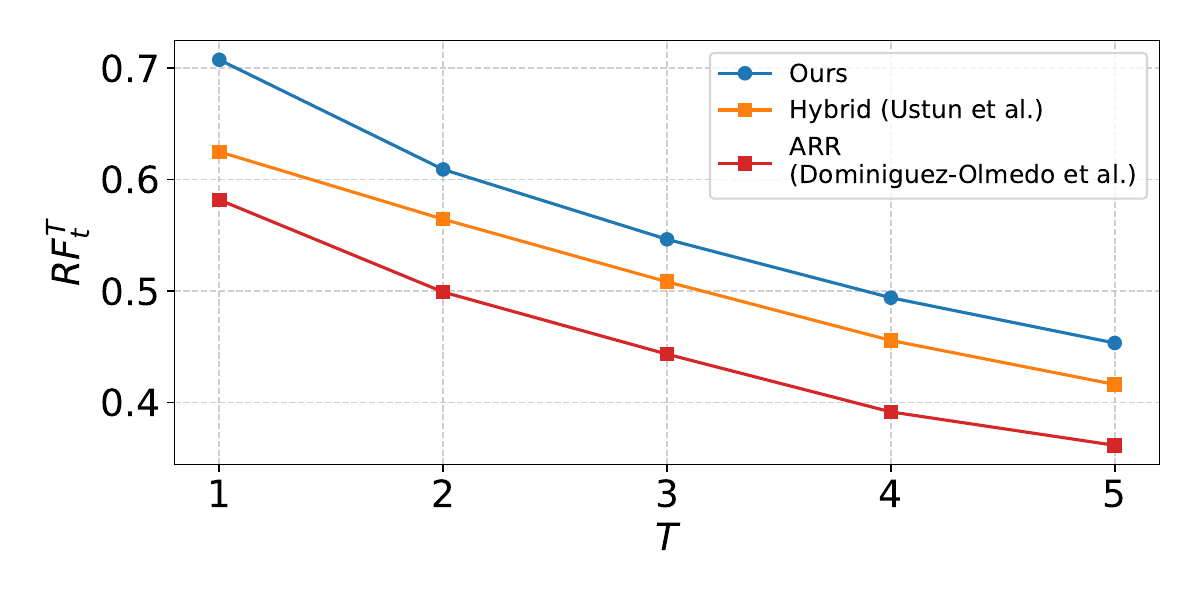}
    \caption{Recourse Feasibility $\text{RF}_t^T$, for a fixed value of Recourse Reliability $\text{RR}_t^T$ ($\approx0.95$), and $\beta=0.05$, varying $T \in [1,5]$, for our method, the hybrid (based on Ustun's approach), and ARR.}
    \label{fig:portionImplementingvsT}

\end{wrapfigure}

\namedpar{Impact of the time horizon $T$} Comparing the top and bottom plots in Figure~\ref{fig:paretoGrid}, we observe that the value of $T$ strongly affects the validity-feasibility trade-off: achieving high validity requires policies with lower feasibility as $T$ increases.
Guaranteeing recourse over a longer horizon imposes a more stringent requirement, forcing the agent to recommend more challenging feature changes. Figure~\ref{fig:portionImplementingvsT} further highlights this phenomenon, by plotting the average Recourse Feasibility $\text{RF}_t^T$, fixing $\text{RR}_t^T=0.95$ and $\beta = 0.05$, for $T \in [1,5]$. As noticed, feasibility must decrease to guarantee large reliability over an increasing time horizon.

\begin{figure}[htbp]
    \centering
    \begin{subfigure}[b]{0.49\textwidth}
        \includegraphics[width=\textwidth]{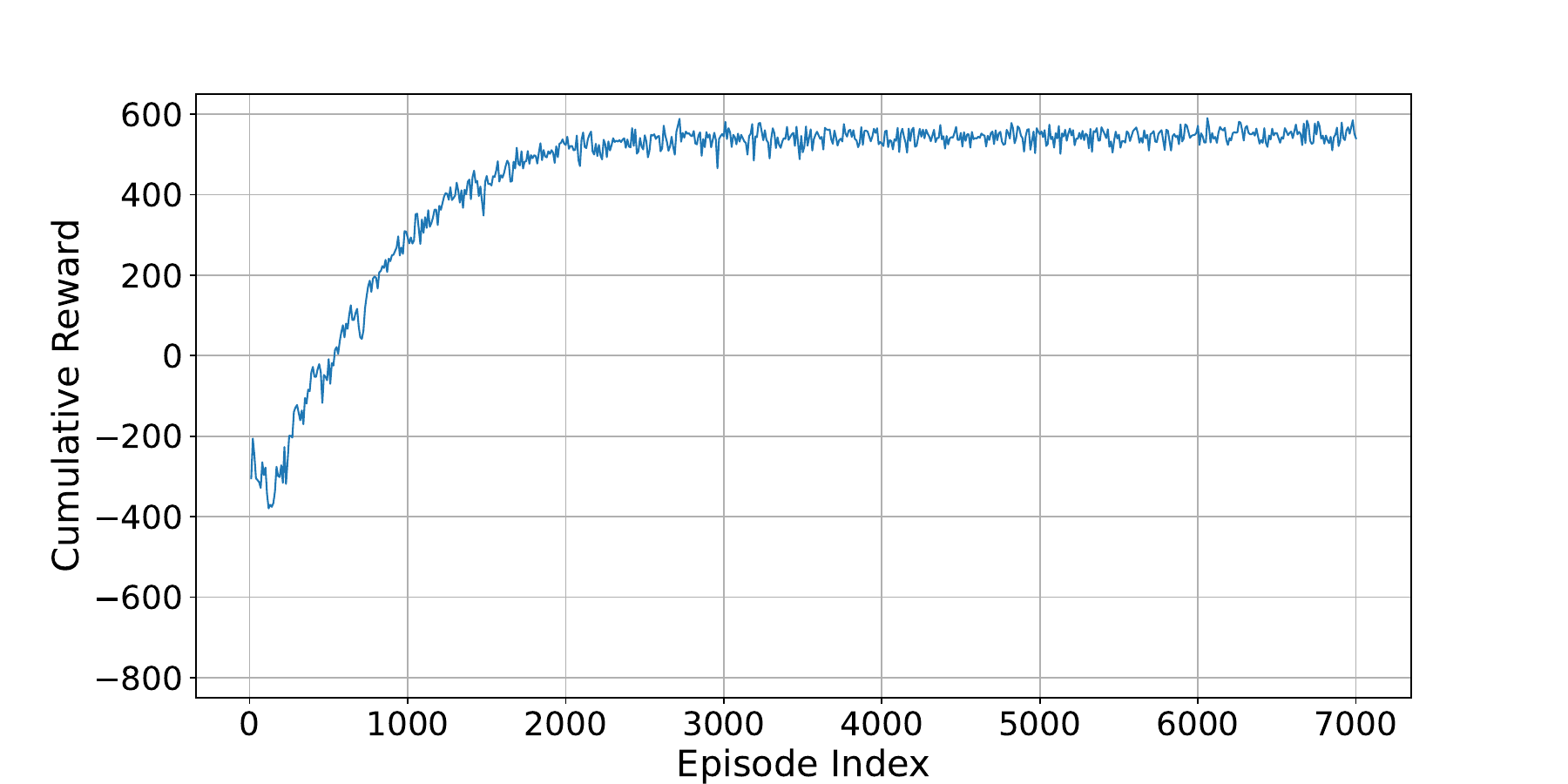}
        \caption{$T=1$}
        \label{fig:convergenceT1}
    \end{subfigure}
    \hfill
    \begin{subfigure}[b]{0.49\textwidth}
        \includegraphics[width=\textwidth]{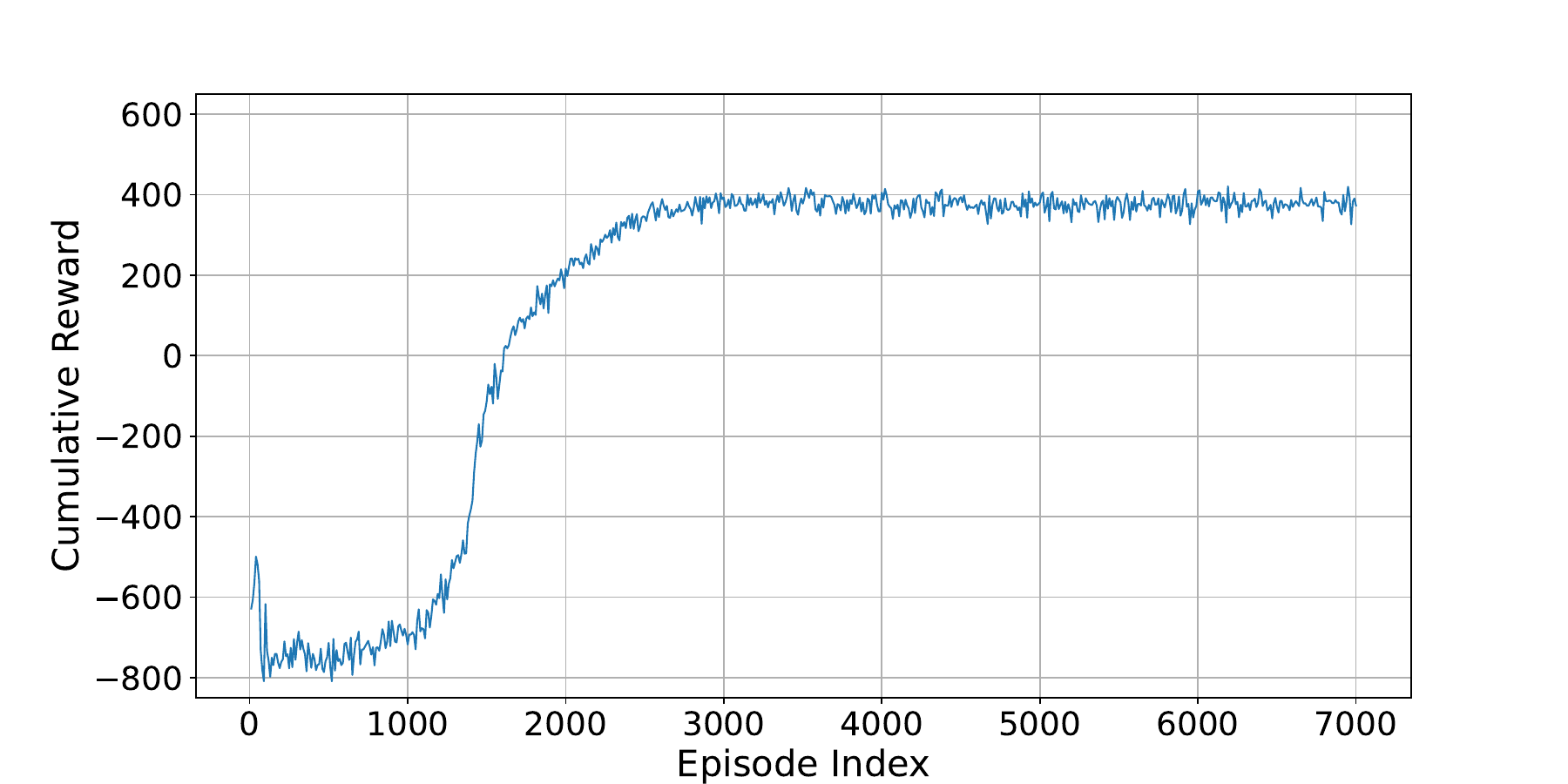}
        \caption{$T=5$}
        \label{fig:convergence_T5}
    \end{subfigure}
    
    \caption{
    Convergence curves in two identical settings ($\beta = 0.01$, $\alpha = 7$, $\tau = 5$), comparing $T \in \{1, 5\}$. The y-axis shows the average cumulative reward (smoothed over ten episodes), and the x-axis denotes the episode index.}

    \label{fig:convergenceTime}
\end{figure}

One additional challenge of a longer time horizon is slower convergence, highlighted in Figure~\ref{fig:convergenceTime}. 
It presents the convergence curves of two predictor agents trained under identical conditions ($\beta = 0.01$, $\alpha = 7$, $\tau = 5$), but with different planning horizons: $T=1$ and $T=5$. Each point represents the cumulative reward averaged over the previous ten episodes. 

For $T=1$, the reward begins to increase after a few hundred episodes and converges to a final value of $\approx 600$ after about $2000$ steps.
In contrast, for $T=5$, the reward starts improving only after roughly $1000$ episodes and reaches a final value of $\approx 400$ after around $3000$ steps. This behavior shows that the agent requires substantially more exploration when validity must be guaranteed over a longer horizon, since the task is more complex.

\namedpar{Impact of the setting difficulty $\beta$}
Comparing the left and right panels of Figure~\ref{fig:paretoGrid} shows that the value of $\beta$ strongly shapes the attainable trade-off between $\text{RF}_t^T$ and $\text{RR}_t^T$. 
We recall that $\beta$ scales the probability of successfully implementing feature modifications: higher values correspond to higher probabilities of success, while lower values make modifications more difficult. In both scenarios, to ensure high reliability, the agent recommends relatively high target scores, that push reapplying candidates above new applicants. For large $\beta$, this strategy has a moderate negative impact on $\text{RF}_t^T$, since even challenging modifications remain feasible. In contrast, for low $\beta$, the same strategy yields a much sharper trade-off, as many candidates are unable to realize the recommended changes. This analysis reveals another intrinsic limitation of recourse in resource-constrained environments. When the means for improvement are inherently difficult (low $\beta$), it is challenging to devise recommendations that are both likely to be implemented and sufficient to guarantee a positive outcome. Consequently, practitioners must carefully prioritize among these desiderata.

Recourse based on Wachter and DiCE highlights the same trends, as shown in Appendix~\ref{app:results}.

\namedpar{Comparison of baseline, hybrid, ARR, and our approach}
Across all subplots in Figure~\ref{fig:paretoGrid}, the baseline approach (green) attains low reliability ($\text{RR}_t^T \approx 0.4$) while favoring feasibility. This outcome reflects the limitations of simply using the last observed decision threshold as the target score instead of tailoring it to an evolving environment. 
In contrast, our predictor $\mu$ can be plugged into any recourse recommender, such as Ustun, explicitly governing this trade-off (orange). These results highlight the advantage of an RL-based predictor over a simplistic fixed threshold policy. 
Comparing ARR (red) to the baseline (green), we observe that robustifying the target score based on a fixed parameter $\varepsilon$ improves recourse reliability, at the cost of reduced feasibility. However, the hybrid method (orange) Pareto-dominates ARR in all four settings, particularly in high-reliability regimes ($RR_t$ above the high reliability threshold). This demonstrates that environment-aware target adjustments outperform naive approaches that ignore the candidates' behavior. A more in-depth analysis of this is provided in Appendix~\ref{app:dominiguez}.


Figure~\ref{fig:paretoGrid} also compares our method (blue) with the hybrid approach (orange). Our approach achieves Pareto optimality across all four settings. The key distinction lies in the recommendation strategy: while Ustun selects recommendations based solely on minimal feature changes, our policy $\phi$ explicitly accounts for feature difficulties, prioritizing changes to easier features, resulting in more feasible recourse paths. Hybrid methods based on Wachter and DiCE achieve a performance similar to Ustun, as illustrated in Appendix~\ref{app:results}. Overall, these results highlight the effectiveness of our method in achieving more favorable reliability–feasibility trade-offs in dynamic, multi-step settings.

\section{Discussion and Conclusions}

This paper presents the first solution to the problem of \emph{robust recourse recommendations in competitive, limited-resource settings}. Our approach leverages reinforcement learning to anticipate candidate responses to recommendations and to generate suggestions that jointly maximize feasibility and validity. By adaptively estimating the relative difficulty of modifying each feature, the method prioritizes more accessible changes. Moreover, it supports recourse validity for $T$ time steps, where $T$ is specified by the stakeholder issuing the recommendations.  

While the RL agent effectively learns environment dynamics, real-world deployment may introduce additional complexities. A key drawback arises during the transient learning phase---especially for the recourse recommender---where candidates may receive suboptimal recommendations. This limitation could be mitigated by training this policy on offline data, such as data from previous recourse systems used in the same setting, which record how the candidate population evolved in response to recommendations. 
Exploring such and extension constitutes an important step toward practical deployment and is left for future work.

Moreover, our simulation environment focuses on non-causal recourse, reflecting the work’s emphasis on robustness under competition. The predictor, however, can be paired with any causal recourse method, and the recommender can be trained in environments with underlying causal structures. Additional sources of uncertainty, such as shifts in new applicants' distribution or in the prediction model, could also be modeled (Appendix \ref{app:env_extensions}). Exploring these extensions provides a promising path for validating robustness in more complex, causally grounded settings. 

Lastly, while our framework empirically shows that RL can be leveraged in  competitive recourse settings to balance multiple desiderata, no theoretical guarantees of convergence to such a solution can be established, as the chosen RL method \citep{haarnoja2018softactorcriticoffpolicymaximum} lacks such guarantees (Appendix \ref{app:convergence_analysis}).
Overall, this work establishes a foundation for durable and adaptive recourse under competition, while opening multiple pathways for further research.

\bibliographystyle{plainnat}  
\bibliography{references}  

@article{upadhyay2021towards,
  title={Towards robust and reliable algorithmic recourse},
  author={Upadhyay, Sohini and Joshi, Shalmali and Lakkaraju, Himabindu},
  journal={Advances in Neural Information Processing Systems},
  volume={34},
  pages={16926--16937},
  year={2021}
}

@inproceedings{fonseca2023setting, series={EAAMO ’23},
   title={Setting the Right Expectations: Algorithmic Recourse Over Time},
   volume={14},
   url={http://dx.doi.org/10.1145/3617694.3623251},
   DOI={10.1145/3617694.3623251},
   booktitle={Equity and Access in Algorithms, Mechanisms, and Optimization},
   publisher={ACM},
   author={Fonseca, João and Bell, Andrew and Abrate, Carlo and Bonchi, Francesco and Stoyanovich, Julia},
   year={2023},
   month=oct, pages={1–11},
   collection={EAAMO ’23} }

@inproceedings{bell2024game, author = {Bell, Andrew and Fonseca, Joao and Stoyanovich, Julia}, title = {The Game Of Recourse: Simulating Algorithmic Recourse over Time to Improve Its Reliability and Fairness}, year = {2024}, isbn = {9798400704222}, publisher = {Association for Computing Machinery}, address = {New York, NY, USA}, url = {https://doi.org/10.1145/3626246.3654742}, doi = {10.1145/3626246.3654742}, booktitle = {Companion of the 2024 International Conference on Management of Data}, pages = {464–467}, numpages = {4}, location = {Santiago AA, Chile}, series = {SIGMOD/PODS '24} }

@inproceedings{
pawelczyk2022trade,
title={On the Trade-Off between Actionable Explanations and the Right to be Forgotten},
author={Martin Pawelczyk and Tobias Leemann and Asia Biega and Gjergji Kasneci},
booktitle={The Eleventh International Conference on Learning Representations },
year={2023},
url={https://openreview.net/forum?id=HWt4BBZjVW}
}

@inproceedings{Altmeyer_2023,
   title={Endogenous Macrodynamics in Algorithmic Recourse},
   volume={12},
   url={http://dx.doi.org/10.1109/SaTML54575.2023.00036},
   DOI={10.1109/satml54575.2023.00036},
   booktitle={2023 IEEE Conference on Secure and Trustworthy Machine Learning (SaTML)},
   publisher={IEEE},
   author={Altmeyer, Patrick and Angela, Giovan and Buszydlik, Aleksander and Dobiczek, Karol and van Deursen, Arie and Liem, Cynthia C. S.},
   year={2023},
   month=feb, pages={418–431} }

@article{wachter2018counterfactualexplanationsopeningblack,
  title        = {Counterfactual Explanations without Opening the Black Box: Automated Decisions and the {GDPR}},
  author       = {Wachter, Sandra and Mittelstadt, Brent and Russell, Chris},
  journal      = {Harvard Journal of Law \& Technology},
  volume       = {31},
  number       = {2},
  pages        = {841--887},
  year         = {2017},
  doi          = {10.2139/ssrn.3063289},
 url          ={https://papers.ssrn.com/sol3/papers.cfm?abstract_id=3063289}
}

@inproceedings{karimi2020algorithmicrecoursecounterfactualexplanations,
author = {Karimi, Amir-Hossein and Sch\"{o}lkopf, Bernhard and Valera, Isabel},
title = {Algorithmic Recourse: from Counterfactual Explanations to Interventions},
year = {2021},
isbn = {9781450383097},
publisher = {Association for Computing Machinery},
address = {New York, NY, USA},
url = {https://doi.org/10.1145/3442188.3445899},
doi = {10.1145/3442188.3445899},
booktitle = {Proceedings of the 2021 ACM Conference on Fairness, Accountability, and Transparency},
pages = {353–362},
numpages = {10},
location = {Virtual Event, Canada},
series = {FAccT '21}
}

@inproceedings{ustun2019actionable, series={FAT* ’19},
   title={Actionable Recourse in Linear Classification},
   url={http://dx.doi.org/10.1145/3287560.3287566},
   DOI={10.1145/3287560.3287566},
   booktitle={Proceedings of the Conference on Fairness, Accountability, and Transparency},
   publisher={ACM},
   author={Ustun, Berk and Spangher, Alexander and Liu, Yang},
   year={2019},
   month=jan, pages={10–19},
   collection={FAT* ’19} }

@InProceedings{haarnoja2018softactorcriticoffpolicymaximum,
  title = 	 {Soft Actor-Critic: Off-Policy Maximum Entropy Deep Reinforcement Learning with a Stochastic Actor},
  author =       {Haarnoja, Tuomas and Zhou, Aurick and Abbeel, Pieter and Levine, Sergey},
  booktitle = 	 {Proceedings of the 35th International Conference on Machine Learning},
  pages = 	 {1861--1870},
  year = 	 {2018},
  editor = 	 {Dy, Jennifer and Krause, Andreas},
  volume = 	 {80},
  series = 	 {Proceedings of Machine Learning Research},
  month = 	 {10--15 Jul},
  publisher =    {PMLR},
  url = 	 {https://proceedings.mlr.press/v80/haarnoja18b.html}
}

@InProceedings{dandl2020multi,
author="Dandl, Susanne
and Molnar, Christoph
and Binder, Martin
and Bischl, Bernd",
editor="B{\"a}ck, Thomas
and Preuss, Mike
and Deutz, Andr{\'e}
and Wang, Hao
and Doerr, Carola
and Emmerich, Michael
and Trautmann, Heike",
title="Multi-Objective Counterfactual Explanations",
booktitle="Parallel Problem Solving from Nature -- PPSN XVI",
year="2020",
publisher="Springer International Publishing",
address="Cham",
pages="448--469",
isbn="978-3-030-58112-1",
  doi       = {10.1007/978-3-030-58112-1_31},
  url       = {https://doi.org/10.1007/978-3-030-58112-1_31}
}

@inproceedings{
pawelczyk2023probabilistically,
title={Probabilistically Robust Recourse: Navigating the Trade-offs between Costs and Robustness in Algorithmic Recourse},
author={Martin Pawelczyk and Teresa Datta and Johan Van den Heuvel and Gjergji Kasneci and Himabindu Lakkaraju},
booktitle={The Eleventh International Conference on Learning Representations },
year={2023},
url={https://openreview.net/forum?id=sC-PmTsiTB}
}

@inproceedings{
cheon2024feature,
title={Feature Responsiveness Scores: Model-Agnostic Explanations for Recourse},
author={Seung Hyun Cheon and Anneke Wernerfelt and Sorelle Friedler and Berk Ustun},
booktitle={The Thirteenth International Conference on Learning Representations},
year={2025},
url={https://openreview.net/forum?id=wsWCVrH9dv}
}

@InProceedings{dominguez2022adversarial,
  title = 	 {On the Adversarial Robustness of Causal Algorithmic Recourse},
  author =       {Dominguez-Olmedo, Ricardo and Karimi, Amir H and Sch{\"o}lkopf, Bernhard},
  booktitle = 	 {Proceedings of the 39th International Conference on Machine Learning},
  pages = 	 {5324--5342},
  year = 	 {2022},
  editor = 	 {Chaudhuri, Kamalika and Jegelka, Stefanie and Song, Le and Szepesvari, Csaba and Niu, Gang and Sabato, Sivan},
  volume = 	 {162},
  series = 	 {Proceedings of Machine Learning Research},
  month = 	 {17--23 Jul},
  publisher =    {PMLR},
  url = 	 {https://proceedings.mlr.press/v162/dominguez-olmedo22a.html}
}

@inproceedings{de2025time,
author = {De Toni, Giovanni and Teso, Stefano and Lepri, Bruno and Passerini, Andrea},
title = {Time Can Invalidate Algorithmic Recourse},
year = {2025},
isbn = {9798400714825},
publisher = {Association for Computing Machinery},
address = {New York, NY, USA},
url = {https://doi.org/10.1145/3715275.3732008},
doi = {10.1145/3715275.3732008},
booktitle = {Proceedings of the 2025 ACM Conference on Fairness, Accountability, and Transparency},
pages = {89–107},
numpages = {19},
location = {
},
series = {FAccT '25}
}

@InProceedings{beretta2023importance,
  author    = "Beretta, Isacco and Cinquini, Martina",
  editor    = "Longo, Luca",
  title     = "The Importance of Time in Causal Algorithmic Recourse",
  booktitle = "Explainable Artificial Intelligence",
  year      = "2023",
  publisher = "Springer Nature Switzerland",
  address   = "Cham",
  pages     = "283--298",
  isbn      = "978-3-031-44064-9",
  doi       = "10.1007/978-3-031-44064-9_16",
  url       = "https://doi.org/10.1007/978-3-031-44064-9_16"
}

@article{karimi2022survey,
author = {Karimi, Amir-Hossein and Barthe, Gilles and Sch\"{o}lkopf, Bernhard and Valera, Isabel},
title = {A Survey of Algorithmic Recourse: Contrastive Explanations and Consequential Recommendations},
year = {2022},
issue_date = {May 2023},
publisher = {Association for Computing Machinery},
address = {New York, NY, USA},
volume = {55},
number = {5},
issn = {0360-0300},
url = {https://doi.org/10.1145/3527848},
doi = {10.1145/3527848},
journal = {ACM Comput. Surv.},
month = dec,
articleno = {95},
numpages = {29},
  doi          = {10.1145/3527848},
  url          = {https://dl.acm.org/doi/10.1145/3527848}
}

@article{de2022personalized,
  title        = {Personalized Algorithmic Recourse with Preference Elicitation},
  author       = {De Toni, Giovanni and Viappiani, Paolo and Teso, Stefano and Lepri, Bruno and Passerini, Andrea},
  journal      = {Transactions on Machine Learning Research},
  year         = {2024},
  url          = {https://openreview.net/forum?id=sh6N4KuDLX}
}

@misc{yang2025robustmodelevolutionalgorithmic,
  author       = {Hao-Tsung Yang and Jie Gao and Bo-Yi Liu and Zhi-Xuan Liu},
  title        = {Towards Robust Model Evolution with Algorithmic Recourse},
  year         = {2025},
  archivePrefix= {arXiv},
  eprint       = {2503.09658},
  primaryClass = {cs.LG},
  url          = {https://arxiv.org/abs/2503.09658},
  note         = {arXiv preprint}
}

@misc{kayastha2024learning,
      title={Learning-Augmented Robust Algorithmic Recourse}, 
      author={Kshitij Kayastha and Vasilis Gkatzelis and Shahin Jabbari},
      year={2024},
      eprint={2410.01580},
      archivePrefix={arXiv},
      primaryClass={cs.LG},
      url={https://arxiv.org/abs/2410.01580},
    doi = {10.48550/arXiv.2410.01580},
note         = {arXiv preprint}

}

@inproceedings{majumdar2024carma,
author = {Majumdar, Ayan and Valera, Isabel},
title = {CARMA: A practical framework to generate recommendations for causal algorithmic recourse at scale},
year = {2024},
isbn = {9798400704505},
publisher = {Association for Computing Machinery},
address = {New York, NY, USA},
url = {https://doi.org/10.1145/3630106.3659003},
doi = {10.1145/3630106.3659003},
booktitle = {Proceedings of the 2024 ACM Conference on Fairness, Accountability, and Transparency},
pages = {1745–1762},
numpages = {18},
location = {Rio de Janeiro, Brazil},
series = {FAccT '24}
}

@InProceedings{segal2024better,
author="Segal, Meirav
and George, Anne-Marie
and Yu, Ingrid Chieh
and Dimitrakakis, Christos",
editor="Longo, Luca
and Lapuschkin, Sebastian
and Seifert, Christin",
title="Better Luck Next Time: About Robust Recourse in Binary Allocation Problems",
booktitle="Explainable Artificial Intelligence",
year="2024",
publisher="Springer Nature Switzerland",
address="Cham",
pages="374--394",
isbn="978-3-031-63800-8",
  doi={10.1007/978-3-031-63800-8_19},
  url={https://doi.org/10.1007/978-3-031-63800-8_19}
}

@inproceedings{venkatasubramanian2020philosophical,
author = {Venkatasubramanian, Suresh and Alfano, Mark},
title = {The philosophical basis of algorithmic recourse},
year = {2020},
isbn = {9781450369367},
publisher = {Association for Computing Machinery},
address = {New York, NY, USA},
url = {https://doi.org/10.1145/3351095.3372876},
doi = {10.1145/3351095.3372876},
booktitle = {Proceedings of the 2020 Conference on Fairness, Accountability, and Transparency},
pages = {284–293},
numpages = {10},
location = {Barcelona, Spain},
series = {FAT* '20}
}

@inproceedings{wu2024safear,
author = {Wu, Haochen and Sharma, Shubham and Patra, Sunandita and Gopalakrishnan, Sriram},
title = {SafeAR: safe algorithmic recourse by risk-aware policies},
year = {2024},
isbn = {978-1-57735-887-9},
publisher = {AAAI Press},
url = {https://doi.org/10.1609/aaai.v38i14.29522},
doi = {10.1609/aaai.v38i14.29522},
booktitle = {Proceedings of the Thirty-Eighth AAAI Conference on Artificial Intelligence and Thirty-Sixth Conference on Innovative Applications of Artificial Intelligence and Fourteenth Symposium on Educational Advances in Artificial Intelligence},
articleno = {1774},
numpages = {9},
series = {AAAI'24/IAAI'24/EAAI'24}
}

@inproceedings{konig2023improvement,
author = {K\"{o}nig, Gunnar and Freiesleben, Timo and Grosse-Wentrup, Moritz},
title = {Improvement-focused causal recourse (ICR)},
year = {2023},
isbn = {978-1-57735-880-0},
publisher = {AAAI Press},
url = {https://doi.org/10.1609/aaai.v37i10.26398},
doi = {10.1609/aaai.v37i10.26398},
booktitle = {Proceedings of the Thirty-Seventh AAAI Conference on Artificial Intelligence and Thirty-Fifth Conference on Innovative Applications of Artificial Intelligence and Thirteenth Symposium on Educational Advances in Artificial Intelligence},
articleno = {1329},
numpages = {9},
series = {AAAI'23/IAAI'23/EAAI'23}
}

@inproceedings{
nguyen2023distributionally,
title={Distributionally Robust Recourse Action},
author={Duy Nguyen and Ngoc Bui and Viet Anh Nguyen},
booktitle={The Eleventh International Conference on Learning Representations },
year={2023},
url={https://openreview.net/forum?id=E3ip6qBLF7}
}

@inproceedings{stkepka2025counterfactual,
author = {Stundefinedpka, Ignacy and Stefanowski, Jerzy and Lango, Mateusz},
title = {Counterfactual Explanations with Probabilistic Guarantees on their Robustness to Model Change},
year = {2025},
isbn = {9798400712456},
publisher = {Association for Computing Machinery},
address = {New York, NY, USA},
url = {https://doi.org/10.1145/3690624.3709300},
doi = {10.1145/3690624.3709300},
booktitle = {Proceedings of the 31st ACM SIGKDD Conference on Knowledge Discovery and Data Mining V.1},
pages = {1277–1288},
numpages = {12},
location = {Toronto ON, Canada},
series = {KDD '25}
}

@inproceedings{karimi2020algorithmic,
author = {Karimi, Amir-Hossein and von K\"{u}gelgen, Julius and Sch\"{o}lkopf, Bernhard and Valera, Isabel},
title = {Algorithmic recourse under imperfect causal knowledge: a probabilistic approach},
year = {2020},
isbn = {9781713829546},
publisher = {Curran Associates Inc.},
address = {Red Hook, NY, USA},
booktitle = {Proceedings of the 34th International Conference on Neural Information Processing Systems},
articleno = {23},
numpages = {13},
location = {Vancouver, BC, Canada},
series = {NIPS '20},
  doi       = {10.5555/3495724.3495747},
  url       = {https://dl.acm.org/doi/10.5555/3495724.3495747}
}

@article{xuan2025perfect,
  title        = {Perfect Counterfactuals in Imperfect Worlds: Modelling Noisy Implementation of Actions in Sequential Algorithmic Recourse},
  author       = {Xuan, Y. and Sokol, K. and Sanderson, M. and others},
  journal      = {Machine Learning},
  volume       = {114},
  pages        = {187},
  year         = {2025},
  doi          = {10.1007/s10994-025-06821-1},
  url          = {https://doi.org/10.1007/s10994-025-06821-1}
}

@InProceedings{guyomard2023generating,
author="Guyomard, Victor
and Fessant, Fran{\c{c}}oise
and Guyet, Thomas
and Bouadi, Tassadit
and Termier, Alexandre",
editor="Koutra, Danai
and Plant, Claudia
and Gomez Rodriguez, Manuel
and Baralis, Elena
and Bonchi, Francesco",
title="Generating Robust Counterfactual Explanations",
booktitle="Machine Learning and Knowledge Discovery in Databases: Research Track",
year="2023",
publisher="Springer Nature Switzerland",
address="Cham",
pages="394--409",
  doi       = {10.1007/978-3-031-43418-1_24},
  url       = {https://doi.org/10.1007/978-3-031-43418-1_24},
isbn="978-3-031-43418-1"
}

@inproceedings{kanamori2025algorithmic,
  title        = {Algorithmic Recourse for Long-Term Improvement},
  author       = {Kentaro Kanamori and Ken Kobayashi and Satoshi Hara and Takuya Takagi},
  booktitle    = {Proceedings of the 42nd International Conference on Machine Learning (ICML 2025) Poster Track},
  year         = {2025},
  month        = may,
  url          = {https://openreview.net/forum?id=gmLD0DHaoZ}
}

@article{rasouli2024care,
  title={{CARE}: coherent actionable recourse based on sound counterfactual explanations},
  author={Rasouli, Pardis and Yu, I. Chieh},
  journal={International Journal of Data Science and Analytics},
  volume={17},
  number={1},
  pages={13--38},
  year={2024},
  publisher={Springer},
  doi={10.1007/s41060-022-00365-6},
  url={https://doi.org/10.1007/s41060-022-00365-6}
}

@inproceedings{rawal2020beyond,
 author = {Rawal, Kaivalya and Lakkaraju, Himabindu},
 booktitle = {Advances in Neural Information Processing Systems},
 editor = {H. Larochelle and M. Ranzato and R. Hadsell and M.F. Balcan and H. Lin},
 pages = {12187--12198},
 publisher = {Curran Associates, Inc.},
 title = {Beyond Individualized Recourse: Interpretable and Interactive Summaries of Actionable Recourses},
 url = {https://proceedings.neurips.cc/paper_files/paper/2020/file/8ee7730e97c67473a424ccfeff49ab20-Paper.pdf},
 volume = {33},
 year = {2020}
}

@inproceedings{upadhyay2025counterfactual,
author = {Upadhyay, Sohini and Lakkaraju, Himabindu and Gajos, Krzysztof Z.},
title = {Counterfactual Explanations May Not Be the Best Algorithmic Recourse Approach},
year = {2025},
isbn = {9798400713064},
publisher = {Association for Computing Machinery},
address = {New York, NY, USA},
url = {https://doi.org/10.1145/3708359.3712095},
doi = {10.1145/3708359.3712095},
booktitle = {Proceedings of the 30th International Conference on Intelligent User Interfaces},
pages = {446–462},
numpages = {17},
location = {
},
series = {IUI '25}
}

@misc{rawal2020algorithmic,
      title={Algorithmic Recourse in the Wild: Understanding the Impact of Data and Model Shifts}, 
      author={Kaivalya Rawal and Ece Kamar and Himabindu Lakkaraju},
      year={2021},
      eprint={2012.11788},
      archivePrefix={arXiv},
      primaryClass={cs.LG},
      url={https://arxiv.org/abs/2012.11788},
    doi = {10.48550/arXiv.2012.11788},
note         = {arXiv preprint}

}

@inproceedings{barocas2020hidden,
author = {Barocas, Solon and Selbst, Andrew D. and Raghavan, Manish},
title = {The hidden assumptions behind counterfactual explanations and principal reasons},
year = {2020},
isbn = {9781450369367},
publisher = {Association for Computing Machinery},
address = {New York, NY, USA},
url = {https://doi.org/10.1145/3351095.3372830},
doi = {10.1145/3351095.3372830},
booktitle = {Proceedings of the 2020 Conference on Fairness, Accountability, and Transparency},
pages = {80–89},
numpages = {10},
location = {Barcelona, Spain},
series = {FAT* '20}
}

@inproceedings{mothilal2020explaining, series={FAT* ’20},
   title={Explaining machine learning classifiers through diverse counterfactual explanations},
   url={http://dx.doi.org/10.1145/3351095.3372850},
   DOI={10.1145/3351095.3372850},
   booktitle={Proceedings of the 2020 Conference on Fairness, Accountability, and Transparency},
   publisher={ACM},
   author={Mothilal, Ramaravind K. and Sharma, Amit and Tan, Chenhao},
   year={2020},
   month=jan, pages={607–617},
   collection={FAT* ’20} }

@article{grbic2012factors,
  title     = {Which factors predict the likelihood of reapplying to medical school? An analysis by gender},
  author    = {Grbic, Douglas and Roskovensky, Lindsay Brewer},
  journal   = {Academic Medicine},
  volume    = {87},
  number    = {4},
  pages     = {449--457},
  year      = {2012},
  publisher = {LWW},
  doi       = {10.1097/ACM.0b013e3182494e54},
  url       = {https://pubmed.ncbi.nlm.nih.gov/22361796/}
}

@article{lievens2005retest,
  title={Retest effects in operational selection settings: Development and test of a framework},
  author={Lievens, Filip and Buyse, Tine and Sackett, Paul R},
  journal={Personnel Psychology},
  volume={58},
  number={4},
  pages={981--1007},
  year={2005},
  publisher={Wiley Online Library},
doi = {10.1111/j.1744-6570.2005.00713.x},
url = {https://onlinelibrary.wiley.com/doi/10.1111/j.1744-6570.2005.00713.x}
}

@book{nelsen2006introduction,
  title={An introduction to copulas},
  author={Nelsen, Roger B},
  year={2006},
  publisher={Springer}
}

\newpage
\appendix
\section{Motivating example}
\label{app:example}
To demonstrate the limitations of existing recourse methods, we examine a Ph.D. admission process. Decisions on admissions are supported by a screening system $M(\cdot)$ that evaluates applicants using criteria such as their GPA, educational background, publications, awards, extracurricular activities, English proficiency, and admission test scores. Admission to the next stage is granted to the top $k$ applicants, where $k$ remains a constant value representing the number of seats available annually.

The goal is to provide rejected candidates with actionable recommendations---feature changes likely to lead to future acceptance, such as, for example, ``Upgrade your education from Bachelor's to Master's'', or ``Increase your test score from 65\% to 70\%''. The motivation for this goal is highlighted by~\citet{venkatasubramanian2020philosophical}: recourse is a fundamental right, and people should be empowered to reverse impactful algorithmic decisions through feasible actions.

State-of-the-art methods typically generate recourse by identifying feature changes that bring a rejected applicant's score to the current threshold. However, this approach can fail in competitive settings. For example, in Figure~\ref{fig:RecourseSetting}, at time $t = 1$, two candidates are accepted and two rejected. Recommendations are given to the rejected candidates to reach the threshold score of 0.51. But at $t = 2$, after implementing these changes, a candidate is still rejected, since more than $k=2$ candidates now meet or exceed the previous threshold. This occurred because the recommendation did not account for the increased competition caused by the recourse itself.

This results in wasted effort, financial cost, and loss of trust in the system. The candidate has acted on the recommendation expecting acceptance, only to be denied again. The issue lies in generating overly easy recommendations that too many can follow, leading to more applicants being able to implement them than available slots. To address this, we propose an approach that anticipates population-level responses and selects more robust target scores. The goal is to ensure that only a subset of candidates can reach these targets, guaranteeing acceptance for those who do. At the same time, recommendations must remain \emph{feasible and actionable}.

Moreover, we introduce the concept of \emph{feature-modification difficulty}, a measure of how difficult it is to change a feature, to reflect real situations constraints. For example, to reach a predefined target score, a candidate might either:
\begin{enumerate}
    \item Publish a first-author paper at a top-tier conference, or
    \item Improve English proficiency from B2 to C1 and increase their test score from 65\% to 85\%.
\end{enumerate}
While the second option requires more changes, it may be more preferred, as the first option requires resources that the student may not have, and entails a high level of uncertainty.  Since precise difficulty ratios are rarely known in advance, we propose estimating them by observing candidate behavior over time.

Finally, recommendations must consider long implementation times and reapplication delays. Following \citet{venkatasubramanian2020philosophical}, we argue that recourse should either be permanent or come with an explicit expiration date. Offering candidates recommendations that are only valid for a single time step risks creating a false sense of agency, since they may have no realistic way to implement the required changes within that interval. We adopt the latter option and associate each recommendation with a \emph{validity horizon} $T$, during which the recommendation ensures acceptance. This allows candidates to plan longer-term changes with confidence that their efforts will remain relevant.

In real deployments, practitioners could select $T$ by combining domain knowledge with empirical evidence. Domain expertise can help estimate the typical time required for individuals to implement meaningful changes to key features, allowing stakeholders to choose a horizon that balances recommendation validity with practical feasibility. Additionally, if historical data were available on applicants who received recommendations and later reapplied, one could empirically estimate the distribution of reapplication intervals or the time needed to implement specific changes.

We note that although the responsibility to ensure longer durability may not be as strong as the responsibility to ensure validity (since the latter corresponds more directly to breaking a promise), both are tied to user trust and to the system’s broader accountability. Failures on either front can lead candidates to disregard the recommendations altogether, ultimately rendering the recourse system ineffective.

\section{Simulation environment details}
\label{app:environment}
\subsection{Synthetic dataset generation}
\label{app:dataset}
To train the predictive model $M(\cdot)$ for estimating candidates' qualification levels, we construct a synthetic dataset of $10{,}000$ examples, each characterized by $10$ continuous features. These examples represent past candidates who were either accepted or rejected. Each feature is independently sampled from a normal distribution, with its mean and standard deviation drawn from uniform distributions, to introduce variability across features. All features are subsequently normalized to lie in $[0,1]$.

Labels are designed to reflect subjective and occasionally inconsistent human decision-making. Specifically, a weighted sum of the features is computed using randomly assigned weights sampled from $[0.1,1]$ and normalized to sum to $1$. Gaussian noise with mean $0$ and standard deviation $0.05$ is added to this score. Candidates with a score exceeding $0.5$ are labeled as accepted; all others are labeled as rejected.

The same feature generation procedure is applied to produce candidate populations $\mathcal{I}_0$ at  the beginning of each episode, and new applicants at each time step. In this case, ground-truth labels are not generated, as they are unnecessary for the simulation.

Our design extends previous simulation environments modeling competitive recourse in limited-resource settings \citep{fonseca2023setting, bell2024game}. Prior work considered candidates generated in a 2-dimensional feature space, sampled independently at random, where $x = (x_1, x_2)$ and $x_i \sim \mathcal{N}(\mu = 0.5, \sigma = 0.3)$ for $i = 1,2$. They trained a simple logistic regressor as a classifier, with target variables $y_i$ drawn from a binomial distribution.

We improve this synthetic data generation process by considering $10$ continuous features instead of $2$, with each feature sampled from a distinct Gaussian. Moreover, we train the logistic regressor on a dataset constructed in this way, where the ground-truth target is correlated with the features, while still incorporating noise.

\subsection{Dropout Probability}
\label{app:dropout}
The likelihood of a candidate dropping out depends on two factors: the gap between their current score and the goal score, and the number of previous applications. Intuitively, candidates are more likely to withdraw when they are far from the goal or have already reapplied multiple times.

Formally, let $b_j = \max(0,g - M(X^{\text{F}}[j]))$ denote the distance of candidate $j$'s score from the goal score $g$, and let $q_j$ be the number of reapplications submitted up to time step $t$. The dropout probability is modeled as a function of these variables, with three decay coefficients: $\rho$ (effect of the score gap), $\chi$ (effect of reapplications), and $\omega$ (their interaction).

\begin{equation}
\label{eq:reapply_prob}
p_{\text{dropout}} = 1 - \exp\!\left(-(\rho b_j + \chi q_j + \omega b_j q_j)\right).
\end{equation}

This exponential form ensures that $p_{\text{dropout}}$ increases monotonically with both $b_j$ and $q_j$, approaching $1$ as either grows large. Conversely, when $b_j=0$ and $q_j=0$, the dropout probability is minimized at $p_{\text{dropout}}=0$, corresponding to a candidate already meeting the goal score on their first attempt.

The term inside the exponent, $\rho b_j + \chi q_j + \omega b_j q_j$, can be interpreted as a \emph{discouragement factor}, jointly capturing how performance shortfall and repeated failures contribute to disengagement.


\subsection{Probability of Successful Implementation}
\label{app:success}
For a candidate $j$ with features $X^{\text{F}}[j]$, the probability of successfully implementing a recommended change on feature $i$ depends on:
\begin{itemize}
    \item the amplitude of the recommended change, $|X^{\text{CF},i}[j] - X^{\text{F},i}[j]|$,
    \item the feature modification difficulty $d_i \in [0,1]$,
    \item the target value $X^{\text{CF},i}[j]$, and
    \item the global scaling parameter $\beta$, which controls the overall difficulty of feature changes.
\end{itemize}
We note that the explicit dependence on the target value reflects the intuition that reaching extreme goals is more challenging, even when the starting point is close.

We define the \emph{attainability} of feature $i$ for candidate $j$ as:
\begin{equation}
\label{eq:attainability}
    a_{j,i} = \frac{1}{|X^{\text{CF},i}[j] - X^{\text{F},i}[j]| \cdot X^{\text{CF},i}[j]} - 1.
\end{equation}

Attainability is minimized at $0$ when $|X^{\text{CF},i}[j] - X^{\text{F},i}[j]| = X^{\text{CF},i}[j] = 1$, and diverges to infinity when any of the denominator terms approaches zero. Intuitively, $a_{j,i}$ quantifies the feasibility of implementing a specific feature change.

The probability of success is then modeled as:
\begin{equation}
\label{eq:p_success}
    p_{\text{success}} = 1 - \exp\!\left(-\beta \cdot \frac{a_{j,i}}{d_i}\right),
\end{equation}
where higher $\beta$ increases the likelihood of success across all features. This probability lies in $[0,1]$ and increases monotonically with attainability. Specifically, when $|X^{\text{CF},i} - X^{\text{F},i}| = d_i = X^{\text{CF},i} = 1$, we obtain $p_{\text{success}}=0$, while if any of these terms is zero, the probability approaches $1$.

\subsection{Probability of Reapplying}
\label{app:reapplying}
At each time step, a candidate’s decision to reapply depends on two factors: \emph{self-confidence}---the extent to which they have implemented the recommendation---and \emph{urgency}---the time elapsed since their last application.

We model the reapplication probability as a convex combination of a distance-based base probability and a time-based scaling factor.

The base probability measures the candidate’s alignment with the goal score. For candidate $j$, it is defined as:
\begin{equation}
\label{eq:p_base}
p_{\text{base},j} = \exp\!\left(-\nu \cdot b_j\right),
\end{equation}
where $\nu$ is a decay parameter, and $b_j$ is the distance of the candidate's current score to the goal score, as previously defined.

The time-based factor captures the increasing tendency to reapply as time passes:
\begin{equation}
\label{eq:urgency}
u_j = \frac{t - l_j}{T},
\end{equation}
where $t$ is the current time step, $l_j$ the last application time step, and $T$ the recourse validity horizon.

The final probability of reapplication is:
\begin{equation}
p_{\text{reapply},j} = (1 - u_j) \cdot p_{\text{base},j} + u_j.
\end{equation}

This formulation guarantees that $p_{\text{reapply},j}$ increases monotonically with time and converges to $1$ either when $u_j = 1$ (i.e., after $T$ steps since the last application) or when $p_{\text{base},j}=1$ (i.e., the recommendation has been perfectly implemented).

\subsection{Possible Environment Extensions}
\label{app:env_extensions}
Our simulation environment is intentionally designed to be extensible, allowing richer behavioral dynamics to be incorporated as needed. In this way, future work could capture more complex candidate behaviors and assess the agent's ability to learn an effective policy in the presence of additional sources of noise.

Table~\ref{tab:phenomena} summarizes several phenomena that could be included, together with indicative implementation strategies.

\begin{table}[htbp]
\caption{Potential extensions to the simulation environment and corresponding implementation strategies.}
\centering
\renewcommand{\arraystretch}{1.5}
\begin{tabular}{p{0.4\textwidth}|p{0.5\textwidth}}
\toprule
\textbf{Phenomenon} & \textbf{Implementation strategy} \\
\midrule
\rowcolor{gray!30} Heterogeneous urgency and self-confidence 
& Introduce personalized parameters in $p_{\text{reapply}}$, sampled per candidate. \\

Collective action among candidates 
& Assign a small probability (e.g., $0.03$) to a coordinated \emph{non-engagement} event. If the event occurs, set $p_{\text{success}} = 0$ for every rejected candidate and set $p_{\text{reapply}} = 1$ at the next time step. \\

\rowcolor{gray!30} A priori low trust in the recourse system 
& For each candidate receiving a recommendation, sample a low-probability event (e.g., $0.05$). If the event occurs, set that candidate's $p_{\text{success}} = 0$, regardless of the attainability (for all the time-steps where such candidate reapplies). \\

Exogenous variation in candidate features 
& Sample a low-probability event (e.g., $0.05$). If the event occurs, select a subset of features and shift the mean or standard deviation of their sampling distributions. \\

\rowcolor{gray!30} Exogenous shifts in the predictive model 
& With small probability (e.g., $0.05$), retrain the model on a modified dataset where the weights for computing the ground truth are slightly perturbed. \\
\bottomrule
\end{tabular}
\label{tab:phenomena}
\end{table}

Each phenomenon requires modifications to specific components of the environment. For instance, personalization of candidates' urgency or self-confidence can be reflected by introducing per-candidate parameters in the reapplication model, scaling the exponent in Equation~\ref{eq:p_base} and $u_j$ in Equation~\ref{eq:urgency}.

Collective action can be modeled as a rare global event in which candidates strategically choose not to pursue recommendations to avoid intensifying competition. Under such an event, candidates follow their usual dropout dynamics via $p_{\text{dropout}}$, but all remaining candidates reapply at the first available time step with unchanged features (i.e., $p_{\text{success}}=0$ and $p_{\text{reapply}}=1$).

Even when collective action does not occur, candidates may individually decline to engage due to low trust in the recourse system. These candidates similarly keep their features unchanged and reapply at the next opportunity (i.e., $p_{\text{success}}=0$ and $p_{\text{reapply}}=1$), regardless of the attainability of their recommendations.

Exogenous shifts may arise in both the candidate population and the predictive model $M(\cdot)$. Shifts in candidate features may be implemented by perturbing the sampling distributions of selected features, modifying their means or standard deviations. For model shifts, retraining $M(\cdot)$ on a perturbed version of the synthetic training set (whose ground-truth weights are slightly altered) provides a simple mechanism to modify the relationship between features and outcomes, thereby affecting the model's learned weights.

\section{Reinforcement learning solution details}
\label{app:solution}

\subsection{Feature Difficulties Estimation}
\label{app:feature_difficulties}
To estimate feature difficulties, we assume partial knowledge of the environment---specifically, the parametric form that links feature difficulties to the probability of successfully implementing a recourse action. Without loss of generality, we fix the parameter $\beta$ as known. As indicated in Equation~\ref{eq:p_success}, $\beta$ acts only as a scaling factor on the difficulties, controlling the overall level of difficulty in the simulation. Consequently, if $\beta$ were unknown, it could be absorbed into the difficulty parameters $d_i$ and estimated jointly with them.

Initially, all estimates of feature difficulties are set to $\hat{d}_i^{(0)} = 0.5$, for all features $i$. After each recourse attempt, we observe whether each feature change was successfully applied, for the only candidate in the environment. Let $y^{(t)}_{i} \in \{0, 1\}$ denote this binary outcome (at time $t$), and let $p^{(t)}_{i}$ be the predicted probability of success, based on the current belief on $\hat{d_i}$. We then compute the error signal:
\begin{equation}
err_{i}^{(t)} = (p_{i}^{(t)} - y_{i}^{(t)}) \cdot a_{i}^{(t)},
\end{equation}
which represents the discrepancy between predicted and observed outcomes, scaled by the attainability $a_{i}^{(t)}$ (previously introduced). 

Feature difficulties are updated using a decaying learning rate:
\begin{equation}
\hat{d}_i^{(t+1)} = \text{clip}\Big(\hat{d}_i^{(t)} + \eta_i^{(t)} \cdot err_i^{(t)}, \, 0, \, 1\Big),
\end{equation}
where
\begin{equation}
\eta_i^{(t)} = \frac{\eta_0}{1 + V_i^{(t)}},
\end{equation}
with $\eta_i^{(0)} = 0.05$ as the base learning rate and $V_i^{(t)}$ the number of prior updates to feature $i$. The clipping ensures that updated difficulties remain within $[0,1]$.

This online procedure allows the model to iteratively refine its estimates of feature difficulties based on observed behavioral responses to counterfactual recommendations.

\subsection{Recourse Recommender Training}
\label{app:recommender_training}
The recourse recommender is trained in a simplified environment with a single candidate. The reward penalizes both the error in Equation~\ref{eq:error} and the cost in Equation~\ref{eq:cost_simplified}.  

To facilitate learning, the reward evolves in two phases. During an initial warm-up period, it depends only on the error term, enabling the agent to learn accurate mappings toward predefined goals. Once feature-modification difficulty estimates stabilize, the cost term is introduced. From this point, the agent operates in a constrained RL setting, where it must choose the \emph{lowest-cost} recommendation among those that reach the target.

The combined reward is:
\begin{equation}
\label{eq:cost_and_error}
    r_t = 
    \begin{cases} 
        -\varphi \cdot c_t, & \text{if } e_t \leq \varepsilon, \\
        -\varphi \cdot c_t - \psi \cdot (e_t - \varepsilon), & \text{otherwise},
    \end{cases}
\end{equation}
where $\varepsilon$ is a tolerance threshold, and $\varphi, \psi$ are hyperparameters with $\psi \gg \varphi$.  

Over training, both the difficulty estimates $\mathbf{d}$ and the recommendation policy converge, yielding a recourse recommender capable of producing accurate and low-cost counterfactuals.

\section{Experimental Setup}  
\label{app:experimental_setup}
The first step in our experimental setup is to construct the score-based decision model $M(\cdot)$. We generate a synthetic dataset of $10\,000$ candidates, each described by 10 features and a binary ground-truth label indicating past acceptance or rejection. $M(\cdot)$ is a logistic regression model, trained on this dataset to approximate the ground-truth labels. The model’s probabilistic outputs serve as candidate scores, representing the estimated likelihood of acceptance. 

The same data generation procedure is used to initialize candidate instances for training the policy $\phi$. Training episodes for the recourse recommender span up to 10 time steps and are conducted in two phases. In the first phase, the reward is based solely on prediction error (Equation~\ref{eq:error}), and training runs for 3,000 episodes. In the second phase, the reward incorporates both prediction error and modification cost (Equation~\ref{eq:cost_and_error}), and training continues for an additional 20,000 episodes. The parameters used are $\varepsilon = 0.01$, $\varphi = 10$, and $\psi = 300$.  

The predictor policy is trained on a simulated population initialized with $N = 20$ candidates. At each time step, $k = 9$ candidates are accepted and $m = 10$ new candidates are introduced. The feature difficulties are set to $\mathbf{d} = [0.84,0.15,0.85,0.78,0.25,0.18,0.29,0.83,0.91,0.10]$. Each episode comprises 100 time steps. The reward function for the predictor is defined as:
\begin{equation}
\label{eq:reward_predictor}
\begin{aligned}
R(s_t, a_t) &= \alpha \cdot \left(1 + 0.90 \cdot \log(\text{RR}_t^T)\right) + \tau \cdot \left(1 + 0.90 \cdot \log(\text{RF}_t^T) \right),
\end{aligned}
\end{equation}
where the logarithmic transformation emphasizes the impact of low values of both metrics. The coefficients $\alpha$ and $\tau$ are positive and adjusted across simulations. The predictor is trained for 7,000 time steps. 

\section{Additional Results}
\label{app:results}
\subsection{Recourse Recommender Policy Performance}
\label{app:rec_results}
\begin{wraptable}{r}{0.5\textwidth} 
\caption{Average prediction error and modification cost---computed with respect to the \emph{true} feature difficulties---for our recourse recommender $\phi$ and the comparison approaches. Each method is evaluated under conditions matching the training setting of our recourse recommender, and results are averaged over ten evaluation runs.}
 
\renewcommand{\arraystretch}{1.5}
\centering
\begin{tabular}{lcc}\toprule
\textbf{Method} & \textbf{$e_t$}& \textbf{$c_t$} \\ \midrule
\rowcolor{gray!30} Ours  & $1.9 \times 10^{-3\phantom{6}}$ & $5.9 \times 10^{-2}$ \\
Ustun & $2.2 \times 10^{-16}$ & $3.0 \times 10^{-1}$ \\
\rowcolor{gray!30} Wachter & $2.6 \times 10^{-3\phantom{6}}$ & $2.7 \times 10^{-1}$\\ 
DiCE &$1.6 \times 10^{-2\phantom{6}}$ & $3.6 \times 10^{-1}$\\ 
\bottomrule
\end{tabular}
\label{tab:ErrorAndCost} 
\vspace{-10pt}
\end{wraptable}

The summed absolute error between the true difficulties $\mathbf{d}$ and their estimates $\mathbf{\hat{d}}$ is given by
\begin{equation}
e_{\text{diff}} = \sum_{i=1}^z |d_i - \hat{d}_i|,
\end{equation}
and is approximately $3 \times 10^{-2}$, indicating high fidelity in the difficulty estimation process.

After training the recourse recommender, we assess its performance using the prediction error from Equation~\ref{eq:error} and the \emph{true} modification cost:
\begin{equation}
c_t = \sum_{i=1}^z |x_{t}^{\text{CF},(i)} - x_{t}^{\text{F},(i)}| \cdot d_i,
\end{equation}
where, relative to Equation~\ref{eq:cost_simplified}, the estimated difficulties $\hat{d}_i$ are replaced with their true values $d_i$. Both quantities are averaged over ten evaluation episodes.

Our method achieves an average error of $1.9 \times 10^{-3}$ and an average cost of $5.9 \times 10^{-2}$ (Table~\ref{tab:ErrorAndCost}).

For comparison, we applied the same protocol to Ustun, Wachter, and DiCE. Ustun’s method achieved near-zero error ($e_t = 2.2 \times 10^{-16}$) but incurred a substantially higher cost ($c_t = 3.0 \times 10^{-1}$). Wachter and DiCE obtained errors of the same order as our method but at high costs, similarly to Ustun.

The strong precision of Ustun’s method is expected: it employs integer programming to compute exact minimal changes for achieving the target score in linear models. In contrast, Wachter’s, DiCE’s, and our RL-based approach rely on approximate, gradient-based or learning-based optimization. Consequently, they exhibit slightly higher error values but remain applicable to a broader class of models, unlike Ustun’s approach which is restricted to linear formulations.

These results highlight the effectiveness of the proposed policy in balancing fidelity to the target decision with minimizing modification cost.

\subsection{Comparison with Wachter and DiCE}
\label{app:wachter_dice}
\begin{figure}[ht]
    \centering
    \begin{subfigure}[b]{0.48\linewidth}
        \centering
        \includegraphics[width=\linewidth]{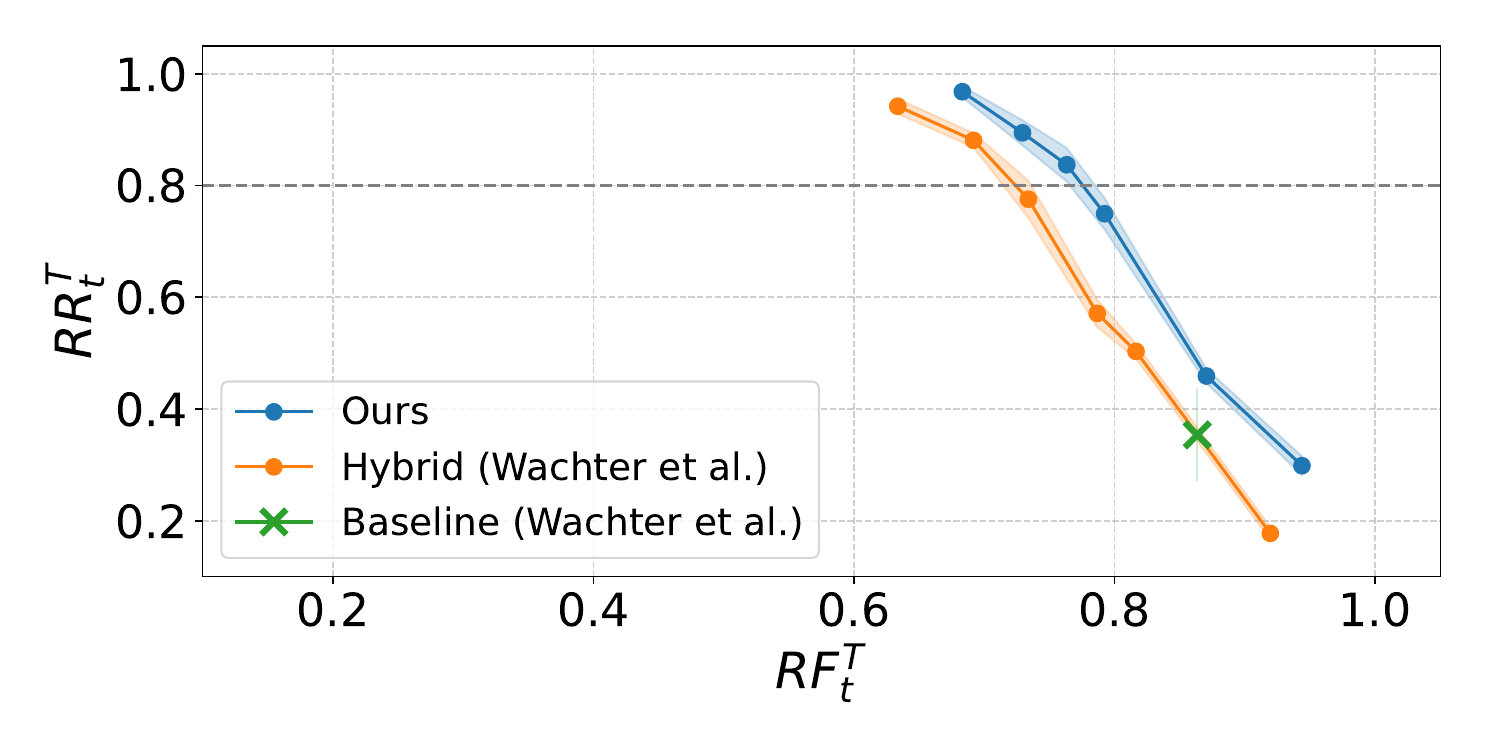}
        \caption{Comparison with Wachter’s approach.}
        \label{fig:paretoT=1EasySetting-Wachter}
    \end{subfigure}
    \hfill
    \begin{subfigure}[b]{0.48\linewidth}
        \centering
        \includegraphics[width=\linewidth]{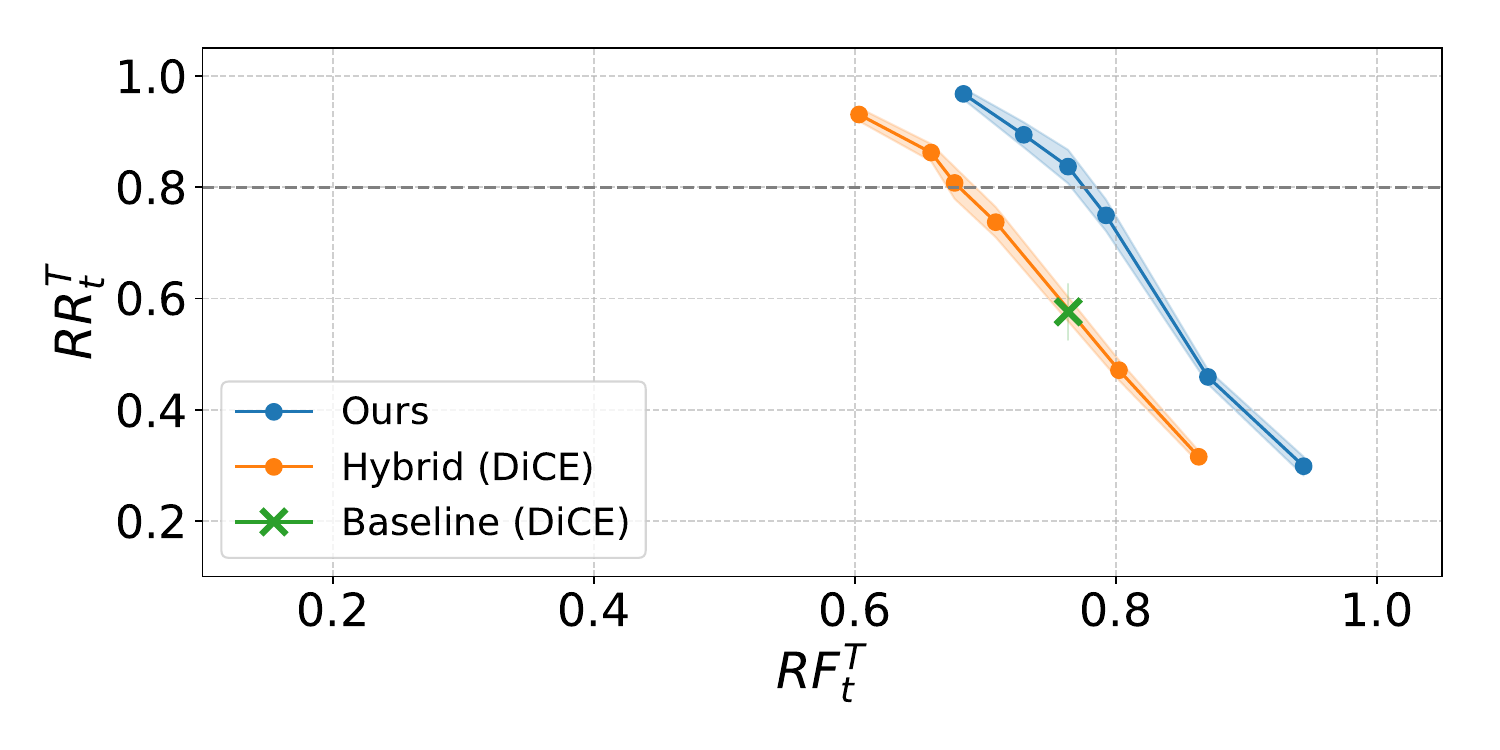}
        \caption{Comparison with DiCE.}
        \label{fig:paretoT=1EasySetting-DiCE}
    \end{subfigure}
    
    \caption{Comparison of Pareto fronts of our method (blue line), the hybrid method based on Wachter's approach and DiCE (orange line), and the baseline method using Wachter's approach and DiCE (green dot), in a setting with $T=1$ and $\beta=0.05$. The Pareto fronts plot the Recourse Reliability $\text{RR}_t^T$ (averaged over ten evaluation episodes) against the Recourse Feasibility $\text{RF}_t^T$ (also averaged over ten evaluation episodes).}
    \label{fig:paretoT=1EasySetting}
\end{figure}

Figure~\ref{fig:paretoT=1EasySetting-Wachter} compares our method with Wachter in a setting with $\beta=0.05$ and $T=1$, while Figure~\ref{fig:paretoT=1EasySetting-DiCE} analogously compares DiCE. As observed previously, the baseline achieves low values of reliability ($\approx 0.4$ for Wachter and $\approx 0.6$ for DiCE), prioritizing feasibility. 
On the other hand, the hybrid variant provides greater control over the trade-off between feasibility ($\text{RF}_t^T$) and validity ($\text{RR}_t^T$), achieving high validity ($\text{RR}_t^T \approx 0.95$) while maintaining feasible recommendations ($\text{RF}_t^T \approx 0.60$), in both cases. 
Overall, the Pareto fronts of the hybrid variants closely match that shown in Figure~\ref{fig:paretoT1EasySetting}, related to Ustun's approach. Our method remains Pareto-optimal, identifying more attainable paths to reach a target score.

\begin{table}[htbp]
\centering
\renewcommand{\arraystretch}{1.5}

\begin{minipage}{0.48\textwidth}
\caption{Comparison of Gini indices. Results are averaged over ten episodes and reported for two settings ($T=1$, $T=5$). All methods are matched on average $\text{RR}_t^T \approx 0.95$ and $\beta=0.05$.}
\centering
\begin{tabular}{lcc}\toprule
& \multicolumn{2}{c}{\textbf{Gini Index}} \\ 
\cmidrule(lr){2-3}
\textbf{Method} & $T=1$ & $T=5$ \\ \midrule
\rowcolor{gray!30} Ours  & $3.2\times10^{-3\phantom{6}}$ & $3.0\times10^{-3}$ \\
Ustun & $2.3\times10^{-16}$ & $2.3\times10^{-16}$ \\
\rowcolor{gray!30} Wachter & $1.8 \times 10^{-4\phantom{6}}$ & $1.8 \times 10^{-4}$ \\ 
DiCE & $1.3 \times 10^{-2\phantom{6}}$ & $1.2 \times 10^{-2}$ \\ 
\bottomrule
\end{tabular}
\label{tab:GiniCost1}
\end{minipage}
\hfill
\begin{minipage}{0.48\textwidth}
\caption{Recourse feasibility ($\text{RF}_t^T$), averaged over ten evaluation episodes, for a fixed recourse reliability ($\text{RR}_t^T \approx 0.95$) and $T=1$, across different values of $\beta$. 
}
\centering
\begin{tabular}{lcc}\toprule
& \multicolumn{2}{c}{\textbf{Recourse Feasibility $\text{RF}_t^T$}} \\ 
\cmidrule(lr){2-3}
\textbf{Method} & $\beta=0.05$ & $\beta=0.01$ \\ \midrule
\rowcolor{gray!30} Ours  & $0.71 \pm 0.01$ & $0.30 \pm 0.02$ \\
Ustun & $0.63 \pm 0.03$ & $0.26 \pm 0.02$ \\
\rowcolor{gray!30} Wachter & $0.61 \pm 0.02$ & $0.30 \pm 0.02$ \\ 
DiCE & $0.55 \pm 0.02$ & $0.25 \pm 0.03$ \\ 
\bottomrule
\end{tabular}
\label{tab:RFvsBeta}
\end{minipage}

\end{table}
\subsection{Analysis of the Gini Index of Each Recourse Recommender}
\label{app:gini}
We evaluate the average Gini index and recommendation cost of our policy $\phi$ as well as the methods by Ustun, Wachter, and DiCE, when paired with our predictor $\mu$. The evaluation considers both $T=1$ and $T=5$. For each recourse recommender, we train a dedicated predictor, ensuring comparability by selecting models that achieve an average Recourse Reliability, over ten evaluation episodes, of approximately $0.95$. The Gini index (Table~\ref{tab:GiniCost1}), defined in Equation~\ref{eq:gini}, is computed over ten evaluation episodes.

The results indicate that varying $T$ has no substantial effect on the Gini index. As expected, Ustun’s method yields extremely low values ($\approx 10^{-16}$), reflecting near-perfect equity. The other methods achieve higher but still reasonably low values. This behavior aligns with our earlier discussion (Section~\ref{sec:solution}): a recourse recommender that more precisely maps to a predefined score produces lower dispersion in target scores, and hence a lower Gini index. Accordingly, the observed indices are strongly correlated with the average errors reported in Table~\ref{tab:ErrorAndCost}. Importantly, while Ustun’s method achieves the greatest precision, our approach delivers equitable recommendations that, as demonstrated in the main text, also attain high feasibility and reliability.

\subsection{Impact of $\beta$ on the Validity-Feasibility Trade-off}
\label{app:beta}
We analyze the effect of $\beta$ on the balance between validity and feasibility. Lower values of $\beta$ correspond to settings in which feature changes are more difficult to implement, making the trade-off between maintaining high validity ($\text{RR}_t^T$) and achieving feasible recourse ($\text{RF}_t^T$) more pronounced.

Table~\ref{tab:RFvsBeta} shows that decreasing $\beta$ substantially worsens the validity-feasibility trade-off. While validity is held fixed ($\text{RR}_t^T \approx 0.95$), feasibility drops sharply: for example, our method’s $\text{RF}_t^T$ falls from $0.707$ at $\beta=0.05$ to $0.365$ at $\beta=0.01$. This highlights that even strong methods face limited options in stringent settings, making the balance between feasible and validity particularly challenging in such settings.

\section{In-depth comparison of our approach, the hybrid (based on Ustun et al.) and ARR (Dominiguez-Olmedo et al.)}
\label{app:dominiguez}
\begin{table}[htbp]
\caption{Recourse feasibility ($\text{RF}_t^T$), averaged over ten evaluation episodes, for a fixed recourse validity ($\text{RR}_t^T \approx 0.95$), varying $T \in \{1,5\}$ and $\beta \in \{0.05, 0.01\}$.}
\centering
\renewcommand{\arraystretch}{1.5}

\begin{tabular}{lcccc}
\toprule
& \multicolumn{2}{c}{\textbf{$T=1$}} & \multicolumn{2}{c}{\textbf{$T=5$}} \\
\cmidrule(lr){2-3} \cmidrule(lr){4-5}
\textbf{Method} & $\beta=0.05$ & $\beta=0.01$ & $\beta=0.05$ & $\beta=0.01$ \\
\midrule
\rowcolor{gray!30} \textbf{Ours} 
& $0.71 \pm 0.01$ & $0.30 \pm 0.02$ & $0.45 \pm 0.01$ & $0.25 \pm 0.01$ \\

\textbf{Hybrid (Ustun et al.)} 
& $0.63 \pm 0.03$ & $0.26 \pm 0.02$ & $0.42 \pm 0.03$ & $0.22 \pm 0.02$ \\

\rowcolor{gray!30} \textbf{ARR (Dominiguez-Olmedo et al.)} 
& $0.57 \pm 0.02$ & $0.22 \pm 0.02$ &  $0.36 \pm 0.02$ & $0.16 \pm 0.01$ \\

\bottomrule
\end{tabular}
\label{tab:comparison_arr}
\end{table}

In Table~\ref{tab:comparison_arr}, we zoom into the results of our approach, the hybrid variant based on \citet{ustun2019actionable}, and the ARR approach \citep{dominguez2022adversarial} in high-reliability regimes (i.e., $RR_t = 0.95$). As shown in the table, our method achieves higher feasibility at the same reliability level than the hybrid approach, by prioritizing changes on features with lower difficulties. Meanwhile, the hybrid approach, which uses our recommender to choose a target score based on the environment's characteristics and candidates' behavior, outperforms the ARR method, whose recommendations consist of a target score derived from a robustified threshold (parameterized by $\varepsilon$) at each time step.

The higher feasibility observed in this case is likely due to the fact that, while ARR robustifies the threshold at every time step, our recommender predicts future competition. As a consequence, in some cases it anticipates a decrease in competition in subsequent time steps and therefore lowers the target score.

Specifically, when examining the behavior of our trained recommender $\mu$ (paired with Ustun) in a short evaluation episode (Table~\ref{tab:evaluation_ustun}) with $\beta = 0.05$ and $T = 1$, we observe that there are time steps in which the agent recommends a target score lower than the most recent threshold ($t = 2$ and $t = 4$). These recommendations still yield high levels of reliability. As a result, the average recourse feasibility $RF_t$ increases, since candidates are only recommended strong changes when necessary.

\begin{table}[htbp]
\caption{Values of the threshold, the subsequent score recommended by the predictor $\mu$, and the observed Recourse Reliability ($RR_t$) and Recourse Feasibility ($RF_t$) at each time step in an 8-step evaluation episode. The quantities $RR_t$ and $RF_t$ are computed relative to the recommendation issued in the previous step. At $t=2$ and $t=4$, the agent recommends a target score lower than the most recently observed threshold. The corresponding values of $RR_t$ and $RF_t$ at the following steps ($t=3$ and $t=5$) are both desirable.
}
\centering
\renewcommand{\arraystretch}{1.5}
\begin{tabular}{lcccccccc}
\toprule
& \multicolumn{8}{c}{\textbf{Time step}} \\
\cmidrule(lr){2-9}
& $t=1$ & $t=2$ & $t=3$ & $t=4$ & $t=5$ & $t=6$ & $t=7$ & $t=8$ \\
\midrule
\rowcolor{gray!30} \textbf{Threshold values}
& 0.46 & 0.63 & 0.60 & 0.63 & 0.56 & 0.57 & 0.59 & 0.54 \\
\textbf{Recommended scores}
& 0.63 & 0.60 & 0.63 & 0.63 & 0.58 & 0.59 & 0.60 & \\
\rowcolor{gray!30} \textbf{Recourse Reliability $RR_t$}
& & 1.0 & 1.0 & 1.0 & 1.0 & 1.0 & 1.0 & 1.0\\
\textbf{Recourse Feasibility $RF_t$}
& & 0.36 & 0.67 & 0.30 & 0.86 & 0.57 & 0.71 & 0.75\\
\bottomrule
\end{tabular}
\label{tab:evaluation_ustun}
\end{table}

\begin{table}[htbp]
\caption{Values of the threshold, the subsequent score recommended by ARR, and the observed Recourse Reliability ($RR_t$) and Recourse Feasibility ($RF_t$) at each time step in an 8-step evaluation episode. The quantities $RR_t$ and $RF_t$ are computed relative to the recommendation issued in the previous step.
}
\centering
\renewcommand{\arraystretch}{1.5}
\begin{tabular}{lcccccccc}
\toprule
& \multicolumn{8}{c}{\textbf{Time step}} \\
\cmidrule(lr){2-9}
& $t=1$ & $t=2$ & $t=3$ & $t=4$ & $t=5$ & $t=6$ & $t=7$ & $t=8$ \\
\midrule
\rowcolor{gray!30} \textbf{Threshold values}
& 0.55 & 0.57 & 0.59 & 0.60 & 0.58 & 0.62 & 0.55 & 0.61 \\
\textbf{Recommended scores}
& 0.61 & 0.63 & 0.65 & 0.66 & 0.64 & 0.68 & 0.61 & \\
\rowcolor{gray!30} \textbf{Recourse Reliability $RR_t$}
&  & 1.0 & 1.0 & 1.0 & 1.0 & 1.0 & 1.0 & 1.0\\
\textbf{Recourse Feasibility $RF_t$}
& &0.36 & 0.44 & 0.33 & 0.71 & 0.71 & 0.43 & 0.50\\
\bottomrule
\end{tabular}
\label{tab:evaluation_arr}
\end{table}

Instead, when examining the behavior of ARR in a short evaluation episode (Table~\ref{tab:evaluation_arr}), we observe that, as expected, the score recommended by ARR is always higher than the most recently observed threshold. As a consequence, the values of recourse feasibility $RF_t$ are generally lower.

\section{Experiment on German}
\label{app:german}
To assess the suitability of our method in real-world settings, we conducted an additional experiment using the German Credit dataset\footnote{\url{https://archive.ics.uci.edu/dataset/522/south+german+credit}}. Although information on how candidates respond to recommendations is not available in this dataset---a limitation relative to literature datasets noted by prior work on recourse in competitive settings \citep{fonseca2023setting}---it can still be used to sample candidates’ initial features, grounding them in a realistic setting. Below, we describe how our experimental setup is adapted for this dataset and present preliminary results.

\subsection{Experimental set-up}
We use the German Credit dataset to train the predictive model $M(\cdot)$ and to sample candidates' initial features. Since the dataset only contains $1\,000$ samples, which is insufficient to both train $M(\cdot)$ and generate a diverse set of candidate profiles for all time steps across episodes, we augment it with synthetic data, generated to be as similar as possible to the original samples.

To preserve the statistical properties of the original dataset, we employ a Gaussian copula approach \citep{nelsen2006introduction}, that captures both the marginal distributions and the correlation structure among features, thus ensuring that synthetic samples closely resemble the original data in terms of both individual feature distributions and inter-feature dependencies.

A key difference between the German dataset and the synthetic dataset that we use in our main experiments is the presence of categorical features. This changes the meaning of feature difficulties: while for continuous features they represent a relative difficulty to implement a feature change, in the case of categorical features they represent the difficulty of switching from a category to another. Thus, for every categorical feature we define a matrix of difficulty parameters, one for every ordered pair of categories. These parameters vary both within the same category (changing the savings from $<100$ to $\geq1000$ is more difficult than changing it to $100\leq x < 1000$), and among different categories (changing the purpose of the loan is easier than changing the savings or the employment). These values represent the difficulty of implementing the corresponding category switch.

The probability of successfully implementing a categorical feature switch $p_{\text{success}}$ is also different from Equation \ref{eq:p_success}, due to these changes. Specifically, we design it as:
\begin{equation}
\label{eq:categorical}
    p_{\text{success}}^{\text{cat}} = (1 - d_{i,j})^{\frac{1}{\beta_{\text{cat}}}}
\end{equation}

where $\beta_{\text{cat}}$ is a setting difficulty parameter, analogous to the parameter $\beta$ previously defined. Lower values of $\beta_{\text{cat}}$ indicate settings where implementing changes to categorical features is generally more difficult. Instead, $d_{i,j}$ represents the difficulty of switching from category $i$ to category $j$.

\subsection{Implementation details}
We train our recommender agent in an environment similar to the one described in Section \ref{sec:simulation} and Appendix \ref{app:environment}, with two key modifications: (i) recommendations on categorical features are constrained to category switches, and (ii) the probability of successfully implementing a category switch follows the formulation defined in Equation \ref{eq:categorical}.

The recommender requires a longer training period compared to the synthetic dataset experiments. This increased complexity likely stems from the challenges inherent in categorical feature recommendations. Specifically, with 17 categorical features spanning 54 distinct categories, the agent must navigate a substantially larger action space of possible feature switches. Moreover, category switches induce non-smooth changes in the credit score, making it difficult to identify counterfactuals that achieve a precise target score. To account for this complexity, we train the recommender for $20\,000$ time-steps in the first phase and $30\,000$ time-steps in the second phase.

The predictor, as in the experiments on synthetic samples, is trained for $7\,000$ time-steps. We use the following hyperparameters: $T=1$, $\beta=0.05$ and $\beta_{\text{cat}}=1.0$.

\subsection{Results}
In Table \ref{tab:German}, we report the results of the preliminary experiment on the German dataset. As can be seen, both our method and the hybrid variant achieve desirable levels of reliability in this more complex setting. On the other hand, as in the scenario with synthetic data, the baseline’s threshold-based policy yields recommendations that too many candidates can satisfy, thereby reducing reliability.

These results showcase the utility of our approach in more complex scenarios, where feature values are drawn from a real dataset and both continuous and categorical features are taken into account.

\begin{table}[htbp]
\centering
\caption{Results for the baseline, the hybrid method (based on Ustun), and our approach in the simulation environment based on German, with $\beta = 0.05$, $\beta_{\text{cat}} = 1.0$, and $T = 1$.}
\renewcommand{\arraystretch}{1.5}
\begin{tabular}{lccc}
\toprule
& \multicolumn{3}{c}{\textbf{Method}} \\
\cmidrule(lr){2-4}
& \textbf{Baseline} & \textbf{Hybrid} & \textbf{Ours} \\
\midrule
\rowcolor{gray!30} \textbf{Recourse Reliability $RR_t$}
& $0.43 \pm 0.05$ & $0.90 \pm 0.04$  & $0.89\pm0.03$ \\
\textbf{Recourse Feasibility $RF_t$} & $0.78 \pm 0.06$ & $0.36 \pm 0.05$ & $0.40 \pm 0.03$ \\
\bottomrule
\end{tabular}
\label{tab:German}
\end{table}

\section{Discussion on convergence guarantees}
\label{app:convergence_analysis}

While our method and the hybrid variants yield desirable solutions in all four settings under consideration ($T \in \{1,5\}$, $\beta \in \{0.05, 0.01\}$), we cannot provide theoretical guarantees that the agent converges to an \emph{optimal} solution. This limitation arises because the reinforcement learning algorithm we employ, Soft Actor Critic (SAC) \citep{haarnoja2018softactorcriticoffpolicymaximum}, does not include such guarantees.

In their paper, the authors provide lemmas and theorems establishing convergence of Soft Policy Iteration, the tabular algorithm SAC is based on. However, these results rely on assumptions that SAC does not satisfy. SAC uses neural networks to approximate policies and value functions, which breaks the assumption of exact or tabular representations, and it operates in continuous action spaces. 
Moreover, SAC incorporates entropy maximization and stochastic actor updates, which further depart from the theoretical setting in which convergence can be proven.

At the same time, deriving strong theoretical bounds in our environment would be extremely difficult, due to the complexity of the environment. Candidates’ behavior is noisy, the system is only partially observable, and the dynamics involve significant stochasticity. Given these complexities, this work is focused on establishing a rigorous empirical framework that demonstrates the practical effectiveness of our approach across diverse experimental conditions.

\section{Use of Large Language Models}
Large language models were used solely to improve the clarity and grammar of the text and to generate the icons of candidates in Figure~\ref{fig:RecourseSetting}. All substantive content was written by the authors; LLMs were applied only for minor phrasing refinements.

\end{document}